\documentclass{article}

\usepackage{microtype}
\usepackage{graphicx}
\usepackage{subcaption} 

\usepackage{titlesec}
\titlespacing*{\section}
{0pt}{2ex plus 1ex minus .2ex}{4.3ex plus .2ex}
\titlespacing*{\subsection}
{0pt}{3ex plus 1ex minus .2ex}{4.3ex plus .2ex}
\usepackage{booktabs} 
\usepackage{multirow}

\usepackage{hyperref}

\usepackage{cleveref}

\usepackage{xcolor} 
\usepackage{algorithm} 
\usepackage[noend]{algpseudocode} 
\newcommand\highlight[1]{\textcolor{blue}{#1}} 
\newcommand\highlightgreen[1]{\textcolor{violet}{#1}} 

\usepackage[accepted]{mlsys2024}

\mlsystitlerunning{Space for Improvement: Exploring the Design Space for Federated Learning in Satellite Constellations}

\begin{document}

\twocolumn[
\mlsystitle{Space for Improvement: Navigating the Design Space for Federated Learning in Satellite Constellations}




\begin{mlsysauthorlist}
\mlsysauthor{Grace Kim}{stan,cam}
\mlsysauthor{Luca Powell}{cam} 
\mlsysauthor{Filip Svoboda}{cam}
\mlsysauthor{Nicholas D.~Lane}{cam}
\end{mlsysauthorlist}

\mlsysaffiliation{stan}{Department of Aeronautics and Astronautics, Stanford University, California, USA}
\mlsysaffiliation{cam}{Department of Computer Science and Technology, University of Cambridge, Cambridge, United Kingdom}

\mlsyscorrespondingauthor{Grace Kim}{gkim65@stanford.edu}
\mlsyscorrespondingauthor{Luca Powell}{lp647@cam.ac.uk}

\mlsyskeywords{Machine Learning, MLSys}

\vskip 0.3in

\begin{abstract}
Space has emerged as an exciting new application area for machine learning, with several missions equipping deep learning capabilities on-board spacecraft. Pre-processing satellite data through on-board training is necessary to address the satellite downlink deficit, as not enough transmission opportunities are available to match the high rates of data generation. To scale this effort across entire constellations, collaborated training in orbit has been enabled through federated learning (FL). 
While current explorations of FL in this context have successfully adapted FL algorithms for scenario-specific constraints, these theoretical FL implementations face several limitations that prevent progress towards real-world deployment. To address this gap, we provide a holistic exploration of the FL in space domain on several fronts. 1) We develop a method for space-ification of existing FL algorithms, evaluated on 2) FLySTacK, our novel satellite constellation design and hardware aware testing platform where we perform rigorous algorithm evaluations. Finally we introduce 3) \textit{AutoFLSat}, a generalized, hierarchical, autonomous FL algorithm for space that provides a 12.5\% to 37.5\% reduction in model training time than leading alternatives.

\end{abstract}
]



\printAffiliationsAndNotice{}  



\section{Introduction}
Recent advancements in small satellite technology have given rise to satellite constellations, which have been identified as an exciting new application area for federated learning (FL) \cite{Lucia2021NanosatConstellationsOpportunitiesChallenges}. This is largely due to satellite data generation rates exceeding downlink opportunities, creating data deficits that grow with constellation size and mission duration. Previous works have tried to address this problem by equipping spacecraft with on board deep learning capabilities, to prioritize high-value data for downlink \cite{Denby2019OrbitalEC, Denby2020OrbitalEdgeComputing, RaVAEn2022UnsupervisedChange}. This has surged a field of FL research to scale this training process across entire constellations, providing utility to the decentralized data stored amongst spacecraft that never have opportunities for downlink to earth \cite{Razmi2022FedSat, Zhai2024FedLEO, elmahallawy2022fedhap, razmi2021onboard, shi2024satellite, chen2021satellite}.

Most existing works have focused on reducing overall convergence times for satellite constellation FL model aggregation \cite{Razmi2022FedSat,razmi2021onboard,Zhai2024FedLEO,elmahallawy2022fedhap}. Compared to traditional FL deployment on edge devices (e.g. smartphones, data clusters), satellites experience a lack of communication windows from periodic client availability to ground stations \cite{razmi2021onboard}. This has prompted many FL in space algorithms to exploit communication protocols specific to satellite constellations such as inter and intra plane links to overcome these constraints \cite{Zhai2024FedLEO, shi2024satellite}. However, none of these works have taken into account how mission specific satellite constellation design and hardware can drastically affect FL performance.
To bridge the gap of FL research into real world deployment, there needs to be a structured review of the most critical satellite constellation design and hardware constraints relevant for FL algorithm design and implementation.

\textbf{Contributions.} In this paper, we holistically explore the design space of FL in LEO constellations with considerations for real-world deployment. The design space comprises three main focus areas, including 1) space-focused FL algorithm development, 2) consideration of both constellation design and satellite hardware for when implementing FL-in-space testing platforms, and 3) rigorous algorithm evaluations under extensive realistic satellite constellation configurations. These are outlined further below.

\textbf{Space Focused FL Algorithms.} In \cref{sec:spaceify}, we provide a \textit{general space-ification framework} for traditional FL algorithms, standardizing the revision of any FL protocol for use under realistic space conditions. Afterwards in \cref{sec:augmentations}, we outline modular augmentations specific to satellite constellation optimizations on the incorporation of scheduling and intra-plane satellite links. These augmentations can be added to any FL algorithm that has undergone space-ification, providing further performance improvements.

Through the insights gained on this space-ification framework, we introduce in \cref{sec:AutoFlSat}, \textit{AutoFLSat}, a fully autonomous satellite FL training framework inspired by principles of hierarchical FL. \textit{AutoFLSat} provides a \textbf{12.5\% to 37.5\%} reduction in model training time compared to leading alternative methods.

\textbf{Constellation Design and Hardware Aware Testing Platforms.} To perform realistic evaluations with satellite orbital trajectories and communication protocols, we introduce in \cref{sec:flystack} \textit{FLySTacK}; this FL in space software pipeline combines satellite access time windows generated by the multi physics application Systems Tool Kit (STK) \cite{rainey_space_2024}, for use in client availability in Flower \cite{beutel_flower_2020}, an open source FL framework. In tandem, to showcase the main hardware constraints for FL applications in space, we develop and introduce in  \cref{sec:FLyCubes} \textit{FLyCubes}, a 1st-generation accessible FL in Space platform suitable for low-cost research missions. 

\textbf{Exploration of Satellite Configuration Space.} Finally, we perform an \textit{extensive exploration of the satellite configuration space} using FLySTacK, with focuses on hardware constraint aware testing. Evaluation is made on the sizes of ground station networks, numbers of clusters in constellations, and density of satellites per cluster against the the space-ified orbital suite of FL algorithms. We supplement this with considerations into hardware constraints of power limitations and data rate in communications with additional experiments in the appendix.


\section{Motivation}
We first discuss current trends in satellite constellations, followed by the scaling problem of downlink deficits. Last, we cover the limitations of existing works exploring FL in satellite constellations.

\begin{figure}[htp]
    \centering
    \includegraphics[width=\linewidth]{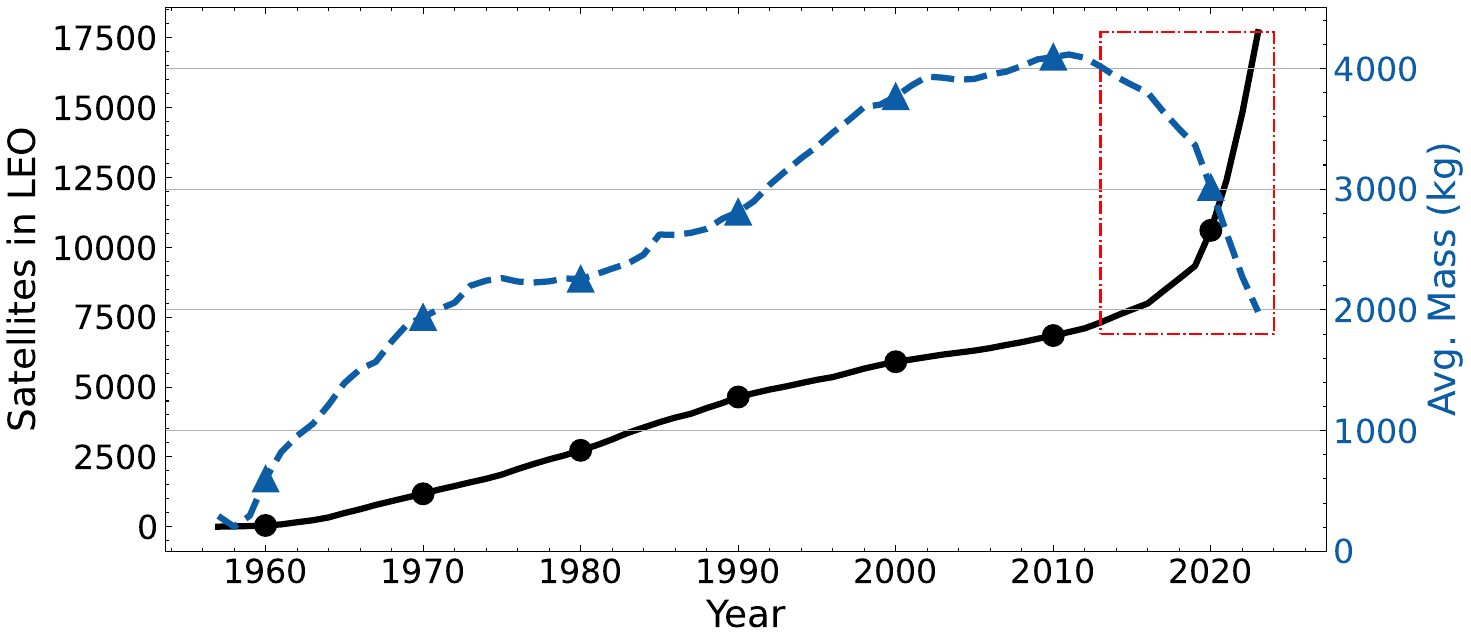}
    \caption{The numbers of satellites in Low Earth Orbit (LEO) and their average mass is displayed. We find a surge in launch cadence of small satellites in the most recent years after 2020, indicating popularity of small satellites within the space industry. Data scraped from DISCOS \cite{DISCOS}.}
    \label{fig:satellite_trends}
\end{figure}

\textbf{Space Industry Trends.}
Small satellites are transforming the space industry with their low development and launch costs, as they can be launched in great numbers as secondary payloads onboard larger spacecraft.
The accelerating launch cadence of increasingly small satellites, illustrated in Figure \ref{fig:satellite_trends}, points to a new paradigm of orbital edge computing (OEC) \cite{Denby2019OrbitalEC, Denby2020OrbitalEdgeComputing}, where small satellites can be deployed at large to form constellations of up to thousands of devices in low-Earth orbit (LEO) \cite{Lucia2021NanosatConstellationsOpportunitiesChallenges}.


\textbf{Downlink Deficits.}
In existing constellations, the rate of data generation often exceeds the data downlink rate, resulting in a `downlink deficit' which grows with each orbit cycle \cite{Lucia2021NanosatConstellationsOpportunitiesChallenges}. The portion of useful data downlinked can be as low as 2\% in some cases \cite{Denby2023Kodan}. The reasons for this can be summarised as follows.

\textit{High-precision Sensors.} Satellites in remote sensing applications contain high-precision sensors onboard, and consequently generate vast amounts of data per orbit cycle. For example, on the Landsat-8's Operational Land Imager (OLI) alone, each 10K hyperspectral image occupies around 1.8GB of data volume, collecting images at a rate of 214,081 per year \cite{LandSat}.
    
\textit{Sparse Ground Station Availability.} Depending on the exact trajectory, the revisit time to ground station can range from 30 minutes to 9 hours \cite{larson_space_1999}. In addition, the rate of ground station network expansion has lagged significantly behind spacecraft launches, as Earth-based facilities can cost up to multi-millions of USD to install \cite{vasisht_l2d2_2021}.
    
\textit{Constrained Downlink Rates.} 
For nanosatellites, downlink rates on VHF and UHF bands typically reach only several KB/s, while earth observation operators achieve slightly higher MB/s rates on L, S, and C bands \cite{info11050258}. Multiple passes are still required to transmit all imagery from one orbit, as higher-frequency X, Ku, and Ka bands (GB/s speeds) are reserved for military, science missions, and internet satellites \cite{Abdollahvand, Maleki}. Bandwidth contention also scales with constellation size.

\textbf{Machine Learning On-Board Spacecraft.} Breaking precedence of the downstream evaluation of satellite data on earth, Machine Learning (ML) methods have been used on board spacecraft to pre-process satellite imagery and address downlink deficits. RaVÆN, an unsupervised change detection model, is an example of one of the first to be deployed and trained directly on the ION SCV004 spacecraft to identify extreme natural events and disasters that should be prioritized for downlink \cite{RaVAEn2022UnsupervisedChange}. 

\textbf{Federated Learning in Space.} 
In addition to onboard inference, motivations for onboard training are apparent in that 1) access to representative training data before mission time is not guaranteed, 2) onboard training provides utility from collected data idle on the spacecraft, and 3) onboard training allows fine-tuning of trained models, with extensions to continual learning. In this regard, federated learning (FL), an established distributed learning paradigm designed to handle decentralised data \cite{mcmahan_communication-efficient_2016}, has been considered a suitable method for collaborative on-board model training across multiple spacecraft \cite{Razmi2022FedSat, Zhai2024FedLEO, elmahallawy2022fedhap, razmi2021onboard, shi2024satellite, chen2021satellite}. FL's scalability for distributed systems, paired with its requirement to process data locally (\textit{on-device}) at each client, makes the method especially applicable to satellite constellations. Since data remains local to each satellite and only model parameters are shared across the constellation, FL effectively accounts for constraints regarding data downlink.

\textbf{Limitations of Existing FL in Space Methods. }
Existing works have predominantly focused on optimising FL algorithms for traits unique to satellite constellations, such as periodic client availability and limited communication windows between satellites and ground stations \cite{Razmi2022FedSat,razmi2021onboard,elmahallawy2022fedhap,shi2024satellite,chen2021satellite}. However, current FL in space methods fail to address key hardware and constellation constraints that limit their application past only ideal case scenarios. These methods assume and apply their strategies onto predefined satellite constellation designs \cite{Zhai2024FedLEO,elmahallawy2022fedhap}, without consideration for generalisation to for different constellation structures or sizes. High data transmission rates, computation capabilities, and power generation is taken for granted \cite{Zhai2024FedLEO,Razmi2022FedSat,razmi2021onboard}, which is not available for all the majority of satellites currently in the Low Earth Orbit (LEO) space. Prior works also overlook the effectiveness of traditional FL algorithms, and how to modify these existing methods for space usage.

\section{Space-ification of FL Algorithms, Building into \textit{AutoFLSat}}\label{sec:algorithms}

Applying FL in space addresses a completely different set of hardware and communication constraints compared to traditional FL on edge devices. 
In \cref{sec:spaceify} we outline a modular approach to standardize the ``space-ification" a traditional FL algorithm, to allow the existing principles to be used under realistic space conditions. We then follow this with augmentation optimizations that are unique to satellite constellations in \cref{sec:augmentations}, focusing on scheduling communications and intra-plane links. Finally, we introduce \textit{AutoFLSat} in \cref{sec:AutoFlSat}, a fully autonomous hierarchical satellite FL training framework developed from our prior algorithm explorations.

\subsection{FL ``Space-ification": Building an Orbital Suite of Satellite Specific FL Algorithms}
\label{sec:spaceify}

By taking into account the space specific constraints fou- 

nd within a satellite constellation, any FL algorithm can undergo a ``space-ification" process for use in a space environment. The formalization of this process was key to informing the development of \textit{AutoFLSat}, a more generally applicable FL algorithm for space in \cref{sec:AutoFlSat}. To outline the process of of how a FL algorithm is ``space-ified", \textit{FedAvg} is taken as an example to showcase the modular nature of the ``space-ify" framework, and serves as a baseline for comparison to alternative algorithms. Similar procedures are taken in \textit{FedProx} and \textit{FedBuff}, algorithms which are uniquely suitable to the communication constraints in satellite constellations, building a standard orbital suite of FL algorithms. The space-ification of \textit{FedProx} and \textit{FedBuff} is detailed in Appendix \ref{appen:spacification_fedprox_fedbuff}.

\begin{algorithm}[!ht]
\caption{\textit{\highlight{FedAvgSat}}. \textit{Base algorithm referenced from the original \textit{FedAvg} paper \cite{mcmahan_communication-efficient_2016}, with modifications for satellite specific operations in \highlight{blue}.} \newline $K$ satellites (clients), $C$ max clients per round, $B$ local minibatch size, $E$ number of local epochs, $\mu$ learning rate, $P_k$ data onboard client $k$, and GS for ground station.}\label{alg:fedAvgSat}
\begin{algorithmic}
\algrenewcommand\algorithmicrequire{\textbf{Main server executes:}}
\algrenewcommand\algorithmicensure{\textbf{ClientUpdate($k$,$w$):}}
\Require 
    \State initialize $w_0$ (global model weights)
\For{each round $t = 0,1,2,...$} 
    \State{$m \leftarrow$ max$(C, K)$} 
    \State{$S_t \leftarrow $\highlight{(first $m$ idle clients that contact a GS)}}
    \State{\texttt{Send} $w_t$ to each $k$ in $S_t$}
    \For{each client $k \in S_t$ \textbf{in parallel}}
        \State{$w^k_t$ =  ClientUpdate $(k,w_t)$}
        \State{\highlight{\textit{Wait} for client $k$ to contact GS again after training}}
        \State{$w^k_{t+1} \leftarrow$ $w^k_t$}
    \EndFor
    \State{$m_t \leftarrow \Sigma_{k \in S_t} n_k$}
    \State{$w_{t+1} \leftarrow \Sigma_{k \in S_t} \frac{n_k}{m_t} w^k_{t+1}$}
\EndFor
\newline
\Ensure \Comment{\textit{Run on client/satellite k}}
\State {\highlight{\textit{Receive}  $w$ from ground station}}
\State{$\beta \leftarrow$ (Split $P_k$ into batches of size $B$)}
\For{ each local epoch $i$ from 1 to $E$}
\For{ batch $b \in \beta$ }
\State{$w \leftarrow w -\mu \nabla l(w;b)$}
\EndFor
\EndFor
\State{\highlight{Wait until reach nearest GS, then return $w$}}
\end{algorithmic}
\end{algorithm}

\textbf{Space-ification Framework, Rebuilding \textit{FedAvg}.} When performing the space-ification of an FL algorithm, there are three main considerations. First, the client selection cannot be randomized because communication windows are too few and every contact must be taken advantage of. This leads to a client selection is based on the first $C$ idle clients that make contact with a ground station. Second, due to the lack of communication windows, the FL round completes once every satellite in the selected clients makes contact again to send back their model parameters. If we assume the main server is not able to check for which satellites have a faster return time for communication (we will discuss this augmentation in \cref{sec:augmentations}) the algorithm must wait until every client included in the round returns their trained weights to complete model aggregation. Finally, the same clients who initially trained the model may not be evaluating the global model after they return the model from training. This is because client selection for evaluation of the aggregate model uses the same protocol as in the training phase in the first consideration. The different clients for training and evaluation will have a more significant effect with larger constellations that have more total clients than the predefined constant $C$. This space-ification applied onto \textit{FedAvg} into \textit{FedAvgSat} is shown in \cref{alg:fedAvgSat}.

By establishing this generalized framework, we have now set up the foundational principles used in \textit{AutoFLSat} for a generalized FL algorithm within the space domain. 

\subsection{Satellite-specific FL Augmentations}
\label{sec:augmentations}

Before presenting the final design of \textit{AutoFLSat}, we describe two augmentations that can be made to further optimize performance of FL algorithms with space-ification. We describe these in isolation for greater clarity, as these augmentations take advantage of satellite specific constraints that are not typically available in a non space domain. These augmentations, scheduling and Intra Satellite Link enabled cluster communication, are inspired by several prior works \cite{elmahallawy2022fedhap} \cite{Razmi2022FedSat} \cite{Zhai2024FedLEO}. 
In this work, both augmentations are expanded for general use, with a framework outlined for modular implementation onto any FL algorithm with space-ification. \textit{AutoFLSat} makes large use of these FL augmentations to reach the performance improvements seen in \cref{sec:autoflsatresults}.

\textbf{Optimizing Deterministic Orbits, Access Scheduler.} Satellite orbits are deterministic, allowing for scheduling of client selections in each round for FL aggregation. This is not typical in traditional FL scenarios. By prioritizing satellites with a shorter combined initial contact and revisit time, duration of rounds are significantly reduced for scenarios with more ground station contact points, and larger constellation sizes. The gains on time optimization for round aggregation can be very significant, as for a LEO satellite with 500km of altitude, the typical access/revisit times of a satellite to a ground station can range from 30 minutes and 9+ hours. For \textit{AutoFLSat}, scheduling becomes incredibly important to effectively perform the method's hierarchical structure of communications.

\begin{figure}[htp]
    \centering
    \includegraphics[width=0.95\linewidth]{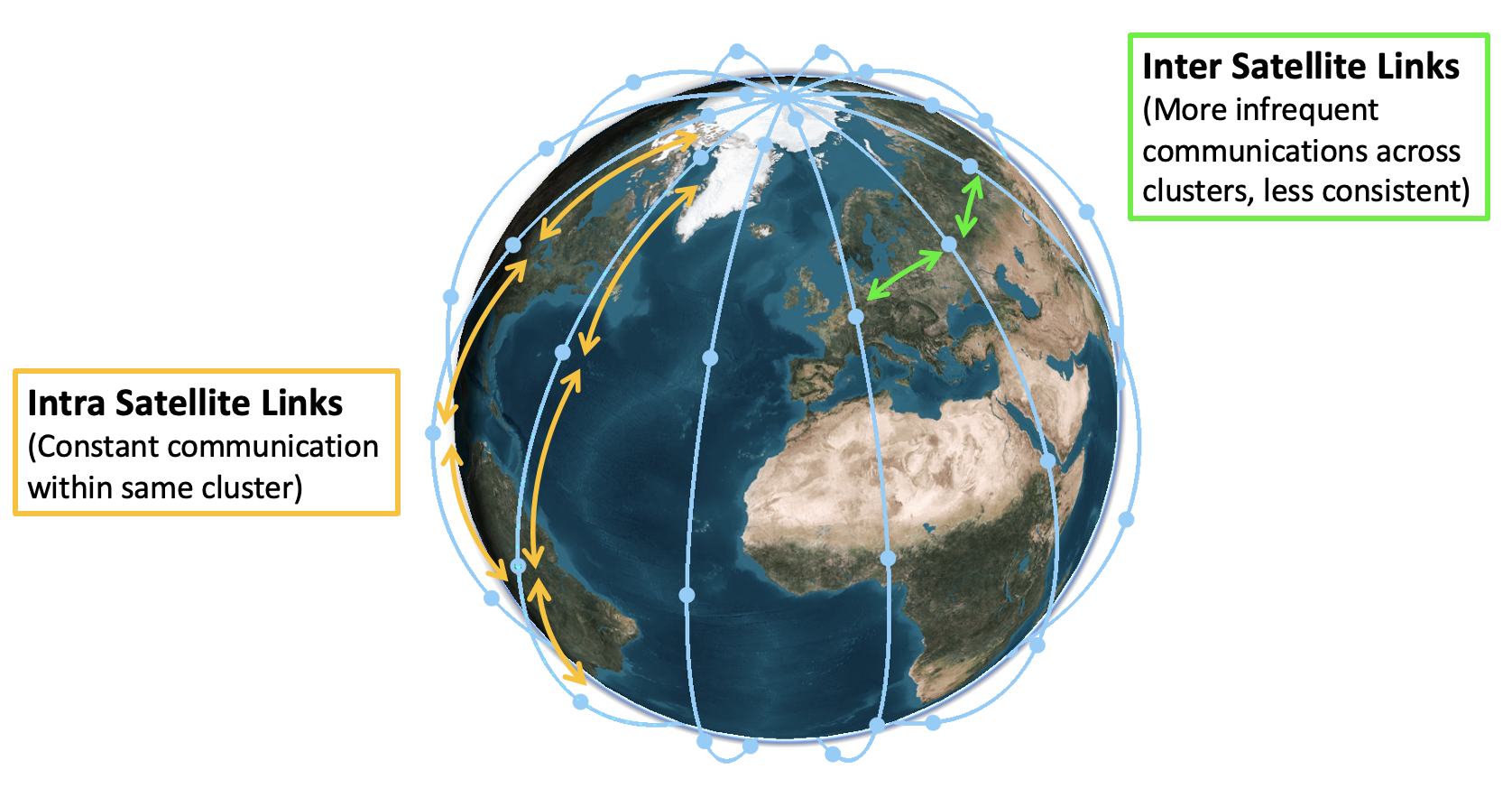}
    \caption{Visual representation of the differences in Intra and Inter Satellite Link (ISL) communication. Intra SL communication is between satellites of the same cluster, and communications are maintained between adjacent satellites at all times. Inter SL communications are much more infrequent, and only occur when different cluster satellites communicate across other orbital planes.}

    \label{fig:ISLs}
\end{figure}

In the appendix, \cref{alg:fedAvg2Sat} provides a brief schematic of the scheduling protocol dubbed \textit{FLSchedule}, which can be applied to any FL algorithm with space-ification in \cref{sec:spaceify}. The general principle only follows two main steps. First, satellite orbits are calculated beforehand in the main server, as well as their access window times to any ground station. Second, prioritization for client selection is made for satellites that have a smaller total initial contact and revisit time to any available ground station.

\textbf{In-Cluster Communication, Intra-Satellite Links.} For large constellations, communication can also be enabled across satellites through Intra or Inter Satellite Links (ISLs). \textit{AutoFLSat} makes use of both Intra and Inter satellite links, but we will discuss only Intra satellite links for this augmentation. Intra Satellite Links refer to communication between satellites in a cluster, as shown in \cref{fig:ISLs}. For satellites spaced closed enough in a cluster, a line of sight to nearby satellites on the same orbital plane can be maintained at all times, eliminating the need for satellites to pass directly over a ground station for data transmission. With proper implementation of ISL scheduling, further reductions are made of satellite idle time or revisit time across a constellation. For Intra SL links, there is a minimum number of satellites needed in a cluster dependent on altitude (e.g. ten satellites on the same cluster for LEO altitude of 500km).

Algorithm \ref{alg:fedAvg3Sat} in the appendix outlines the use of Intra SL communications in a modular addition to a FL algorithm with space-ification and scheduling. Three main considerations are needed to add this augmentation. First, instead of scheduling based only on access windows between satellites and ground stations, cluster specific satellite to satellite communication is also considered. Second, several checks now need to be made to ensure not only the satellite currently training a model can access a ground station, but that any satellite on the cluster can make contact with a ground station. Third, priority for communication would be set to have the \textit{original satellite} send data back to the ground station if both the original and a peer cluster satellite can establish contact with a ground facility.

\subsection{\textit{AutoFLSat}, a Framework for Satellite Constellation Autonomy}
\label{sec:AutoFlSat}

\begin{algorithm}
\caption{\textit{AutoFLSat}. The $K$ satellites (clients) are indexed by $k$, the $G$ ground stations (servers) are indexed by $g$, $C$ clusters, $C_t$ clients in each cluster, $e$ for number of epochs. }\label{alg:fedAvg4Sat}
\begin{algorithmic}
\algrenewcommand\algorithmicrequire{\textbf{On each Satellite Client:}}
\algrenewcommand\algorithmicensure{\textbf{ClientUpdate($k$, $w$, $e$):}}
\algnewcommand\Scheduler{\item[\textbf{InterSLScheduler}$(L_k, L_g, C, t)$]}
\Require 
    
    \\
    \State{Initialize cluster rounds $t$ and $C_t$ clients in cluster}
    \State{Initialize local model $w_0^l$ from $w_0$}
\newline
\For{each cluster round $t = 0,1,2,...$} 

    \State{$e, C_{con} \leftarrow$  \textbf{InterSLScheduler}$(L_k, L_g, C, t)$}
    \State{$w_{t+1}^l \leftarrow$ \textbf{ClientUpdate}$(l,w_t^l,e)$ }\Comment{\textit{All satellites in cluster train for epochs $e$}}
    \newline
    \For{each client $k \in C_t$ } 
    \Comment{\textit{Agg. in cluster}}
         \State{\textbf{Receive} $w_{t+1}^k$ from each $k$ in $C_t$} 
    \EndFor
    \State{$m_t \leftarrow \Sigma_{k \in S_t} n_k$}
    \State{$w_{t+1}^c \leftarrow \Sigma_{k \in S_t} \frac{n_k}{m_t} w^k_{t+1}$} 
    \State{$w_{t+1}^l \leftarrow w_{t+1}^c$}
    \Comment{\textit{Update local model}}
    \newline
    
    \For{each cluster $ \in C$ }  
    \Comment{\textit{Agg. cross clusters}}
    \State{Complete all comms as indicated in $C_{con}$}
         \State{\textbf{Receive} $w_{t+1}^c$ from each cluster $C$} 
    \EndFor
    \State{$m_t \leftarrow \Sigma_{k \in S_t} n_k$}
    \State{$w_{t+1}^C \leftarrow \Sigma_{k \in S_t} \frac{n_k}{m_t} w^c_{t+1}$} 
    \State{$w_{t+1}^l \leftarrow w_{t+1}^C$}
    \Comment\textit{{Update local model from new constellation wide model}}
    \EndFor
    \newline

\Scheduler
\State{Initialize $C_{con}$}
\Comment{\textit{Cluster Connections list}}
\State{$T \leftarrow $ Start time $t$}
\Comment{\textit{Set start time of orbit \textbf{propagation}}}
\newline
\While{ length of $C_{con} < \frac{(C-1)C}{2}$} 
\State{\textbf{Propagate} every sat $k_o$'s orbital trajectories}
\If{$k_o$ can InterSL comm with other sat $k_c$}    
    \State{\textbf{Calculate} access duration length of $k_o$ to $k_c$}
    \If{model params can send in access window}
        \If{first time clusters communicated}
            \State{Add $k_o$ and $k_c$ comms to $C_{con}$}
        \Else
            \If{comms before latest $C_{con}$}
                \State{Update $k_o$ and $k_c$ comms in $C_{con}$}
            \EndIf
        \EndIf
    \EndIf
\EndIf
\EndWhile
\newline
\State{\textbf{Calculate} \# of training epochs $e$ from first and last comms record in $C_{con}$}
\State{\textbf{Return} $e$ and comms record $C_{con}$}
\newline
\Ensure \Comment{\textit{Training on satellite $k$}}
\State{$\beta \leftarrow$ (Split $P_k$ into batches of size $B$)}
\For{ each local epoch $i$ from 1 to $e$}
\For{ batch $b \in \beta$ }
\State{$w \leftarrow w -\mu \nabla l(w;b)$}
\EndFor
\EndFor

\end{algorithmic}
\end{algorithm}

Inspired by the modifications made in \cref{sec:spaceify} and \cref{sec:augmentations}, we introduce \textit{AutoFLSat}, a novel method that makes use of both inter and intra satellite links to develop a completely autonomous federated learning operation, with no need for a central model aggregator. Access to a ground station is needed only at the initialization stage to share the starting global model to one client, which can then disseminated to all clients. 

Influences in hierarchical FL are also found within \textit{AutoFLSat}'s design, with each cluster acting as separate synchronous FL operation, and the aggregation of all clusters acting as another layer of federated aggregation \cite{briggs_federated_2020}. Each cluster can conduct their own synchronous FL operation, as satellites on the same cluster have constant communication to one another through Intra Satellite Links. This means the same number of training epochs can be taken on each cluster's satellite clients without causing delays compared to previous methods. Whenever long enough periods of inter-satellite communication is possible, communications of the global model on each cluster is shared between the communicating satellites. 

Scheduling is used to find the shortest period of time to share models between all clusters. These new global models are then disseminated to all of the respective satellites on the same cluster. No single satellite is designated as the main communication point, so any satellite could receive and disseminate new global models from other clusters at any point in time. The offloading of global models to a ground station can also be performed from any satellite.

Another important note is that the \# of epochs trained on is also dependent on the scheduling taken place to aggregate across clusters. This allows for synchronicity in model training within clusters, but the number of epochs trained on different clusters may have slight differences each round.


\section{FLySTacK, A Platform for Constellation Design and Hardware Aware Constraint Testing}\label{sec:design_platforms_sims_and_hardware}

To properly test the algorithms outlined in \cref{sec:algorithms}, we developed FlySTacK, a novel FL in space software pipeline that combines both realistic satellite modeling and a FL training platform. This is provided as a framework for future researchers to guide further explorations into the FL in space domain. With FLySTacK, we outline several key considerations into the satellite configuration space and hardware constraints tested within our evaluation in \cref{sec:experiments_results}. 

\subsection{FLySTacK - STK x Flwr Simulation Platform}
\label{sec:flystack}

The base of FLySTacK's framework is a combination of the satellite orbital trajectories generated by the multi physics application Systems Tool Kit (STK) \cite{rainey_space_2024}, for use in client availability in Flower \cite{beutel_flower_2020}, an open source FL framework. 
Our framework, FLySTacK, allows FL simulation of a given constellation with a configurable ground station network, constellation topology, and hardware constraints. Using FLySTacK, we performed an extensive exploration into the satellite configuration and hardware space each of our space-ified FL algorithms.

\subsubsection{Satellite Configuration Space} 
For the satellite configuration space, three main parameters were analyzed in depth: the \# of satellite clusters, \# of satellites per cluster, and \# of ground stations available for communications. The combination of these different parameter sweeps allows the setup of several important satellite scenarios for evaluation. These include 1) Single/Multiple Ground Station only communications, 2) Intra Satellite Link communications, and 3) Autonomous Satellite constellation operation. The specific parameters are outlined below.


\textbf{Constellation Parameters.} Inspired from the Doves LEO earth observation imagery satellite constellation from Planet Labs \cite{noauthor_planetscope_2023}, satellites for experiments were set for polar orbits of 90 degrees inclination, to allow the satellite to capture images of all areas on earth with an on-board sensor. These orbits are circular (eccentricity = 0), with an altitude of 500 km above the earth's surface. Depending on the constellation parameters, (i.e. different \# of satellites per cluster) each orbit had the RAAN set accordingly to keep equal spacing among each satellite within a cluster. 

To make use of these polar orbits, the constellation design was set to use a Walker Star constellation as a base. 
Simulations were limited to upward bounds of 100 satellites per constellation, leading to testing configurations with a maximum of 10 clusters, and a maximum of 10 satellites per cluster.

\textbf{Ground Station Placements.} With inspiration from the International Ground Station (IGS) Network \cite{wulder_global_2016-1} which supported the operation of the Landsat 8 and 9 earth observation spacecraft, 13 ground station locations were selected, with a map of the network of ground stations displayed in \cref{fig:groundstations} of the appendix. This constellation network was selected due to its widespread geographic coverage of access points, which would provide more opportunities for communication for all satellites within a constellation. 

\begin{figure*}[!ht]

    \begin{subfigure}{\textwidth}
    \centering
    \includegraphics[trim={7cm 0 7cm 0},clip,scale = 0.25]{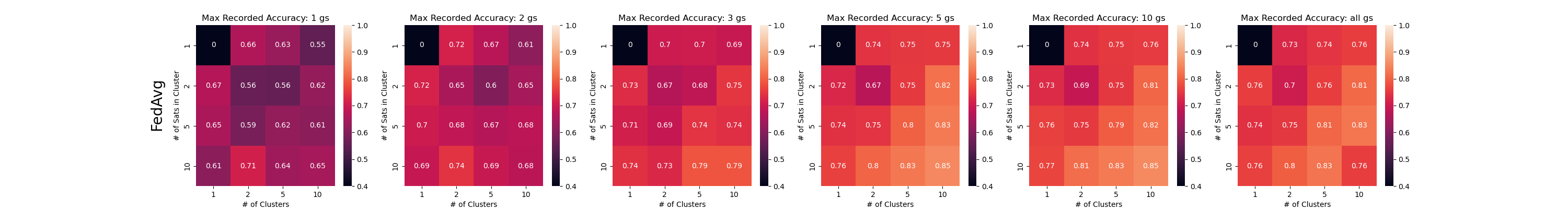}
    \caption{FedAvg with space-ification, accuracy heatmaps}
    \label{fig:fedavg:acc_1}
    \end{subfigure}
    \begin{subfigure}{\textwidth}
    \centering
    \includegraphics[trim={7cm 0 7cm 0},clip,scale = 0.25]{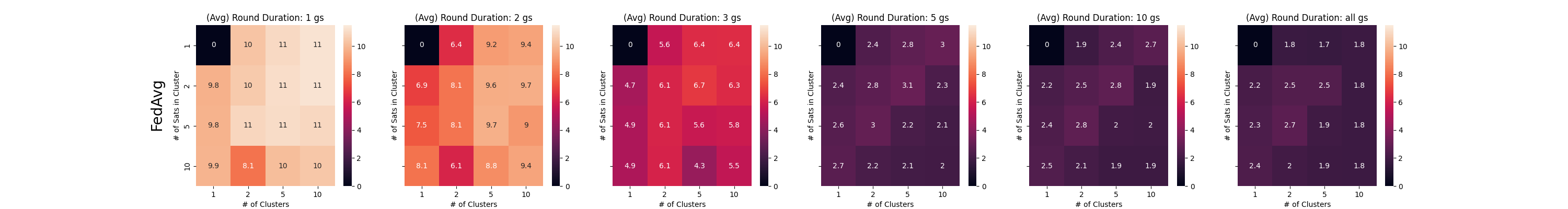}
    \caption{FedAvg with space-ification, round duration heatmaps}
    \label{fedavg10:duration_1}
    \end{subfigure}
    \begin{subfigure}{\textwidth}
    \centering
    \includegraphics[trim={7cm 0 7cm 0},clip,scale = 0.25]{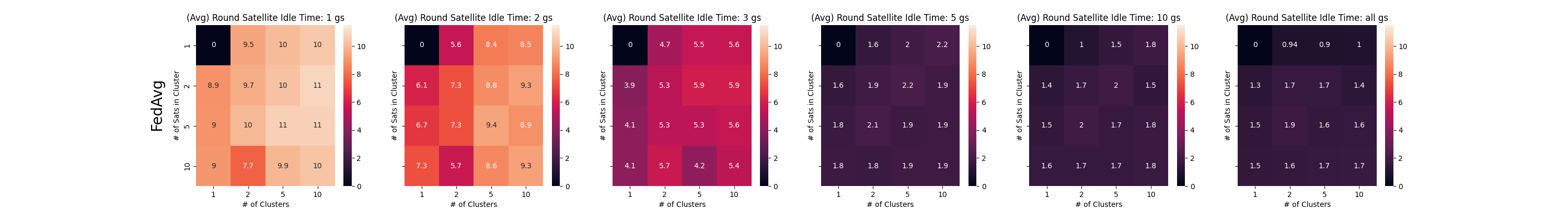}
    \caption{FedAvg with space-ification, idle time heatmaps}
    \label{fedavg10:iavg_1}
    \end{subfigure}
    
    \caption{An example set of heatmaps for FedAvg with space-ification, outlining performances in accuracy, FL round durations, and idle time. Even on the most barebones FL algorithm, we see that with large enough constellation sizes and ground station networks, enough opportunities for access windows can be made to reach convergence. Proper comparisons against performance of other algorithms can be found in the appendix, highlighting the importance of the augmentations on FL algorithm performance. }\label{fig:fedavgheatmap}
\end{figure*}

\subsubsection{Hardware Constraints}
The hardware constraints accounted for in FLySTacK are motivated by discoveries made during the development and testing of FLyCubes (\textbf{F}ederated \textbf{L}earning Read\textbf{y} \textbf{Cube}Sats), our FL-in-space platform prototype which is documented in \cref{sec:FLyCubes}.

\textbf{Orbital Average Power.} The duty cycle of time spent in power modes for different FL tasks - idle, transmitting, training, training+transmitting (full power) can quickly become a constraint depending on power generation and storage capabilities available on each spacecraft. If power generation cannot match the usage of power from various FL tasks, constellations may exhibit delays in transmission of models for FL aggregation as well as interrupted training cycles.

\textbf{Data Rate.} Although low data rates are a large motivator of the need for FL for on board data processing for satellites, low data rates can also become a constraint for satellite constellations that have short contact times with ground stations or ISL links. We provide a short discussion on data rates in relation to inter-satellite link communication in \cref{sec:flycubes_discussion}.

\section{Evaluation}\label{sec:experiments_results}

The results are separated into three main sections. First, \cref{sec:results_param_exploration} discusses and identified the critical design factors tested in the satellite configuration explorations and their consequential effects on accompanying FL methods. These factors led into the key design decisions that established the base framework of \textit{AutoFLSat}. Second, \cref{sec:autoflsatresults} dives into the performance evaluation of \textit{AutoFLSat}, comparing the algorithm against the leading alternatives. 
The autonomous aggregation method provides a 12.5\%-37.5\% reduction in aggregation time compared to other leading alternatives. 

\subsection{Satellite Configuration Space Exploration}\label{sec:results_param_exploration}

To properly evaluate the FL space-ification algorithms in \cref{sec:spaceify}, we assessed the performance of several FL algorithms with space-ification (\textit{FedAvg}, \textit{FedProx}, \textit{FedBuff}, \textit{FedAvgSch}, \textit{FedProxsch}, \textit{FedAvgIntraSL}, \textit{FedProxIntraSL}) with three metrics: 1) Accuracy, 2) FL Round Durations, 3) Satellite Idle Time per Round. Thorough heatmaps of all the different parameter sweeps with these metrics are provided in the appendix, with an example heatmap set for the \textit{FedAvg} space-ification evalutation in \cref{fig:fedavgheatmap}. High level results and conclusions are discussed in this section.

\begin{figure*}[htbp]
    \centering
    \includegraphics[width=\linewidth]{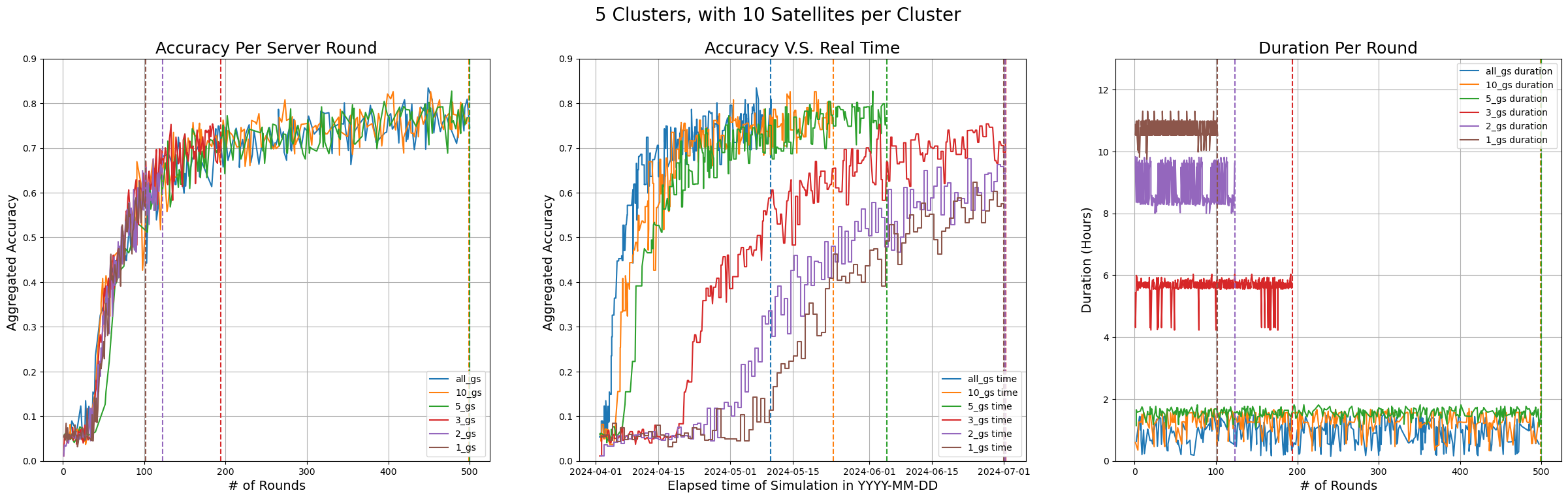}
    \caption{Performance of \texttt{FedAvg} on a satellite constellation with 5 clusters, and 10 satellites per cluster. (Left) Accuracy is plot against server rounds, and it is shown that smaller ground station networks with only 1 or 2 stations can't perform as many aggregation rounds. (Center) In the second plot, accuracy is now plotted against simulation time, which shows simulations with smaller ground station networks struggling to converge even after a 3 month orbital period. (Right) A final plot of duration times for each round is shown, revealing the large jumps in FL round duration time depending on the number of ground stations available to the network.}
    \label{fig:acc_line_fedAvg}
\end{figure*}

\textbf{Experimental Setup.} For the experiments themselves, parameter sweeps were made of the number of clusters, satellites per cluster, and ground stations available for communication. This multi dimensional sweep led to a three dimensional grid of tests for $[1,2,5,10]$ clusters, $[1,2,5,10]$ satellites per cluster, and $[1,2,3,5,10,13]$ ground stations, corresponding to varying stages of satellite constellation sizes in relation to available budget. Within these parameter sweeps, tests were completed for each base FL space-ification algorithm with scheduling and Intra SL enabled communication methodology augmentations. An outline of these experiments can be found in \cref{tab:clflConfigs} of the appendix.

\textbf{Datasets.} For model training, three labeled image datasets were evaluated on: FEMNIST, CIFAR-10, and EUROSAT. FEMNIST and CIFAR-10 were implemented as a performance benchmark to compare against other competing FL in space methodologies, and EUROSAT was selected to provide a real satellite imagery dataset for training.

\subsubsection{Accuracy}
\label{subsec:accuracy}
Before claiming any time based optimizations from an algorithm, a baseline performance of model accuracy must be met. Sweeping over the satellite constellation design parameters, and the FL algorithms with space-ification, the accuracy achieved training on the FEMNIST dataset are displayed in the heatmaps in \cref{fig:accs} in the appendix. An example accuracy heatmap for \textit{FedAvg} with space-ification is shown in \cref{fig:fedavg:acc_1}. Comments on further understanding how to read the heatmap itself is provided in \cref{app:readingheatmaps}. In our findings, we find that even working under strict satellite constellation constraints, \textbf{all FL algorithms were able to
reach at least an 80\% accuracy rate} if provided frequent enough access points to perform aggregation steps. However, \textit{without any augmentations onto traditional FL algorithms with space-ification, reaching performance guarantees in a reasonable time frame is difficult}.


The relationship between time and accuracy becomes clear in \cref{fig:acc_line_fedAvg} analyzing the performance of the \textit{FedAvg} space-ification algorithm. Without scheduling, some constellation configurations couldn't converge past a training accuracy of 60\% over a three month period. However, with scheduling, training accuracy jumped to 80\% as more FL rounds could be performed. Similar time v.s. accuracy graphs with the implementation of \textit{FedAvgSch} is provided in \cref{fig:acc_line_fedAvg2} of the appendix, showcasing
some model time trainings decreasing to 10 days, \textit{a stark $9X$ reduction from the initial 3 month simulation}. By increasing access opportunities through scheduling, intra satellite links, and larger ground station networks, \textbf{previously impractical FL in space timeframes can now be reduced into those within reason.} 

However we note, there are limits to how long much one should optimize the satellite access windows, as enough time must be allotted to complete local computation for each client before aggregation. This led to decreases in performances for specifically \textit{FedProxSch}, prompting the development of \textit{FedProxSchV2}, where a minimum number of epochs needed to be completed before returning parameters to a ground station. 


\subsubsection{FL Round Durations}
\label{sec:roundurr}

In \cref{subsec:accuracy}, tight correlations could be observed between the time allotted to perform training, number of FL rounds completed, and model training accuracy. As FL training across different constellation setups will take the same number of rounds (if given same model and datasets) reducing the time taken for each round will play a key factor in decreasing model convergence times. 
Once again, heatmaps for FL round durations can be found in \cref{fig:base_durations} of the appendix. To compare the duration performances between each FL algorithm in a more concise manner however, a violin plot is made in \cref{fig:summaryDuratinos} also in the appendix. In these explorations, we find that the \textbf{speedups in FL round durations are most significant using larger satellite constellations} as they have numerous access windows available to optimize for in client selection. There are \textit{limits to how much one can optimize time wise over a small network}, which leads into several key satellite constellation design considerations for future FL researchers.


For any optimization to occur over FL round duration times, access window times must be increased through either additional ground stations or ISL communication links. Although ground stations do seem to have the most notable effect in available communication windows, the benefit of adding in ground stations plateaus after a certain number of ground stations incorporated in a network. The experimental setup for these tests suggest that \textit{any more than 5 ground stations introduces redundancy within the system}. To avoid installing more multi-million dollar ground station facilities \cite{vasisht_l2d2_2021}, focusing on increasing cluster specific ISL communication methods may be a more efficient method for large constellations  to increase access window opportunities. 

For satellite constellation designers wishing to leverage ISL communications, it is advised to add more satellites within a cluster rather than increase the number of clusters available. 
Intra SL communication links can only be enabled with a certain number of satellites at different altitudes, which motivates investment into increasing existing clusters rather than building additional ones. Once again, scheduling only becomes useful in reduction in FL round duration times once there is a large enough space of access windows to choose from, which is not often guaranteed with Inter SL communications compared to Intra SL.




\begin{figure*}[htbp]
\centering
\begin{subfigure}{.3\textwidth}
  \centering
  \includegraphics[width=
0.96\linewidth]{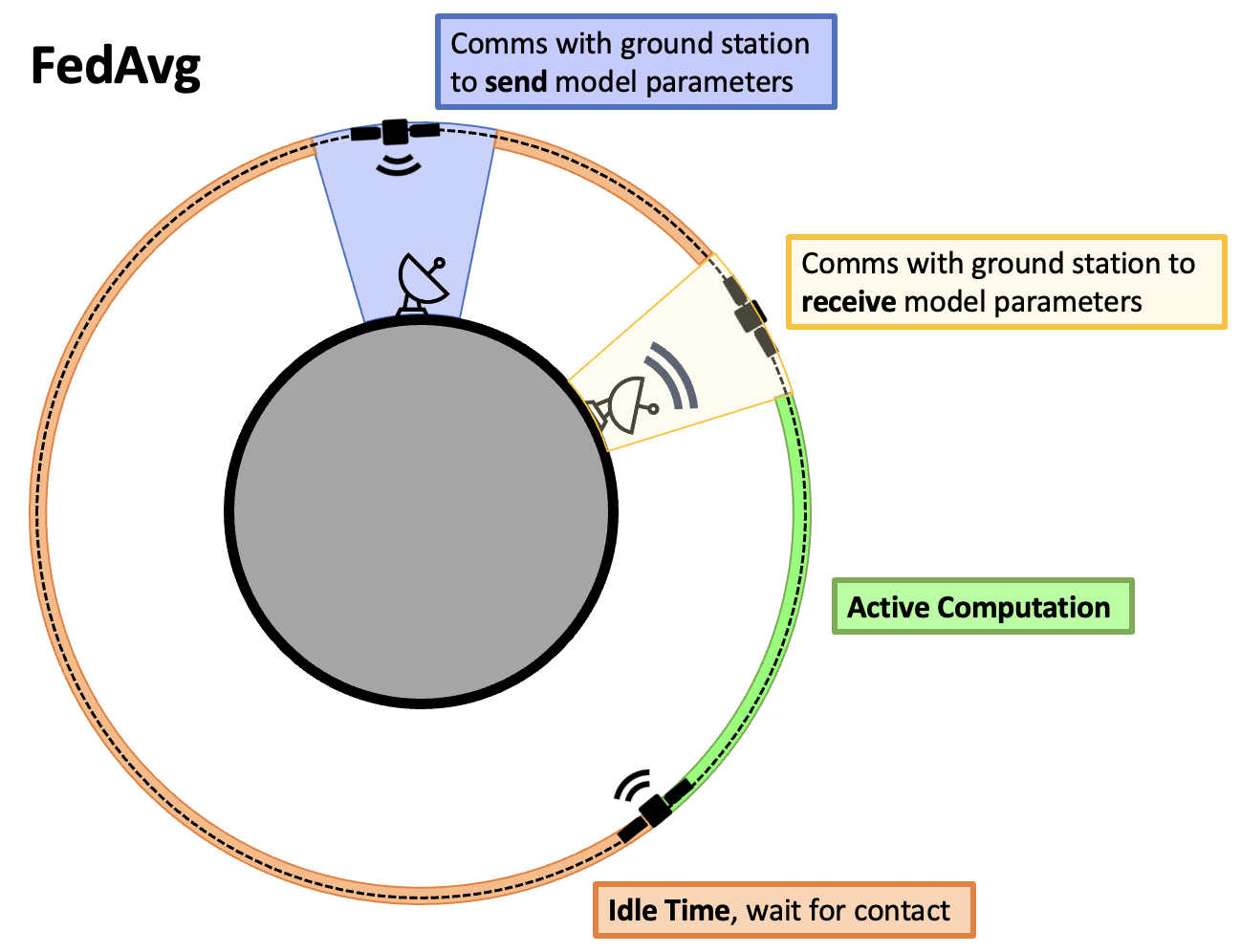}
  \caption{Idle times seen for a satellite in a \textit{FedAvg} FL aggregation setup.}
  \label{fig:sub1}
\end{subfigure}%
\hfill
\begin{subfigure}{.3\textwidth}
  \centering
  \includegraphics[width=\linewidth]{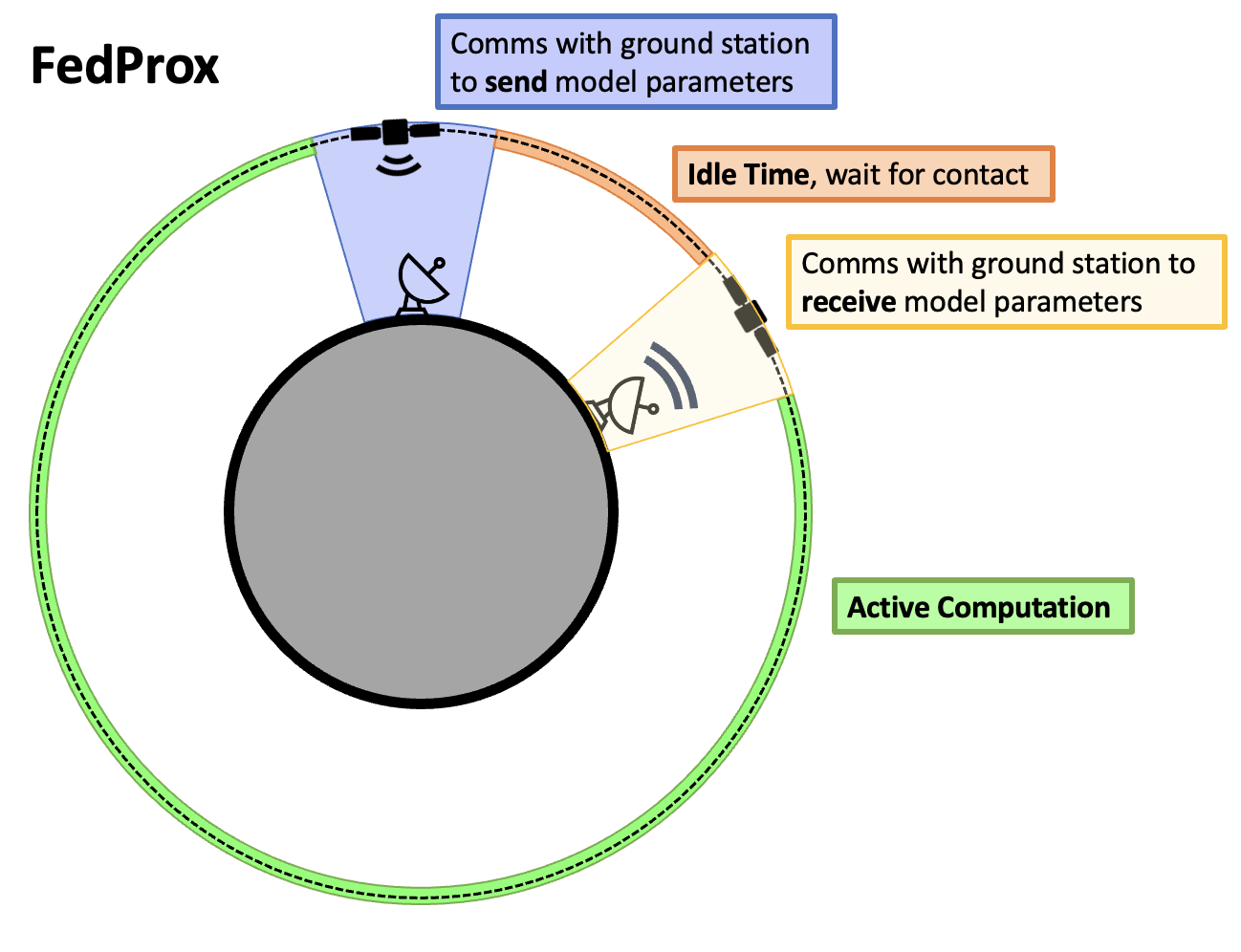}
  \caption{Idle times seen for a satellite in a \textit{FedProx} FL aggregation setup.}
  \label{fig:sub2}
\end{subfigure}
\hfill
\begin{subfigure}{.3\textwidth}
  \centering
  \includegraphics[width=\linewidth]{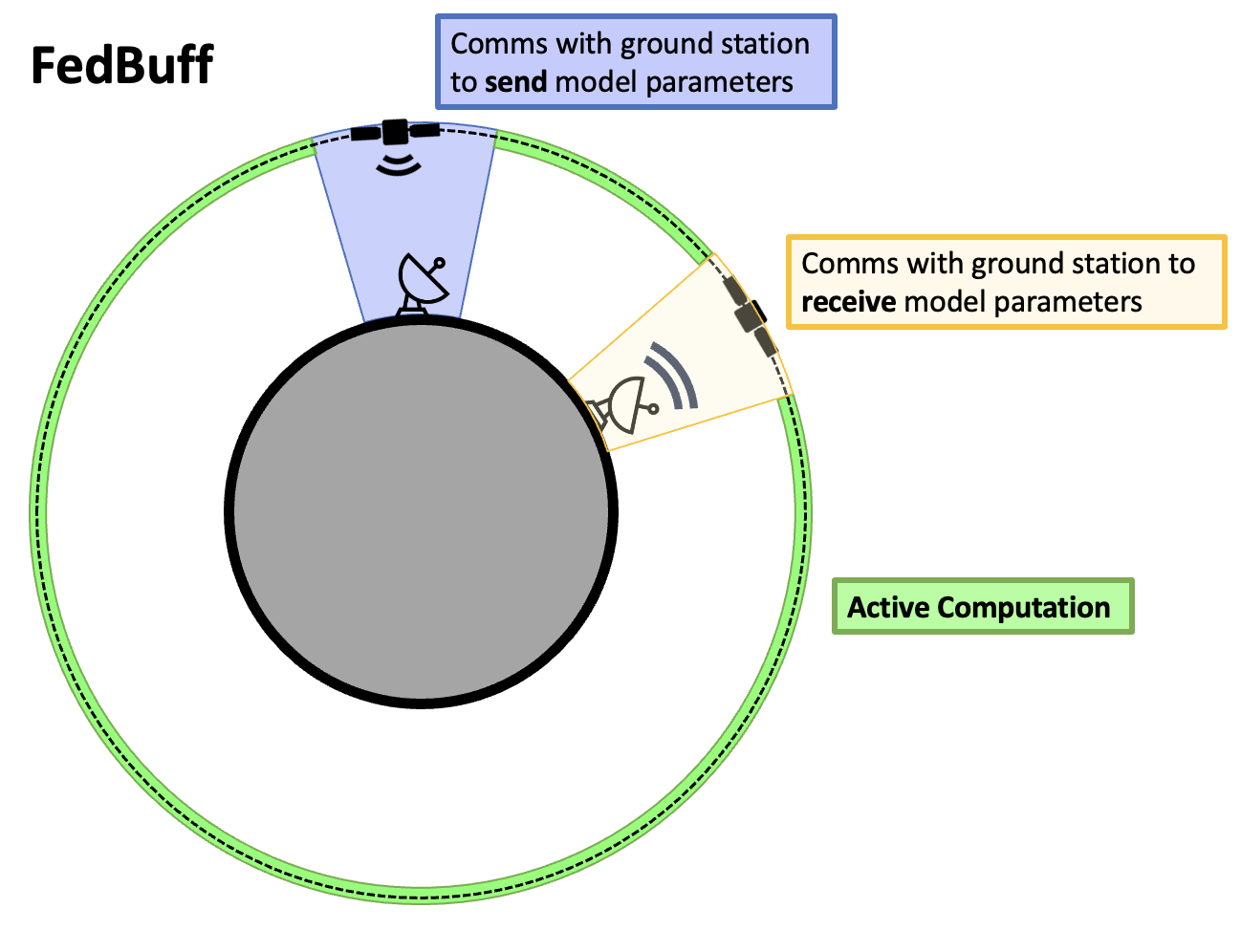}
  \caption{Idle times seen for a satellite in a \textit{FedBuff} FL aggregation setup.}
  \label{fig:sub3}
\end{subfigure}
\caption{Idle times for each aggregation method broken down along an example satellites orbit, with orange representing idle times, blue representing satellite to ground station communication times, yellow for ground station to satellite communication times, and green for active computation time on-board the satellite. \textit{FedBuff} has almost virtually no idle time in comparison to \textit{FedAvg}, which waits both in the model sending and receiving portion of the satellites orbit, and \textit{FedProx} which waits in the model receiving stage of the FL aggregation.}
\label{fig:idledescription}
\end{figure*}

\subsubsection{Satellite Idle Time}
\label{subsec:idletime}
To paint the full picture of how each round is spent on each satellite, the idle time that is accrued specific to each method must be evaluated to understand where the most optimizations in aggregation time can be made. When calculating idle time, any period that is not spent on computation or communication is considered idle. Due to the different communication and aggregation strategies on each FL algorithm, the way idle time accrues on each method widely differs. A more detailed breakdown of the makeup of each FL round duration is explored in \cref{fig:idledescription}.

Similar to the results found in the FL round durations experiments, the number of ground stations played a large role in determining the length of idle times found in each round of the satellite. A similar plateauing effect of reduction in idle time could be found after a certain number of ground stations were added to the network. This however, may be due to the fact that FL round duration length for each scenario automatically caps the maximum length of idle time that can be seen on the satellite. We conclude that \textbf{no matter how much idle time is optimized, more benefit is gained focusing on reducing lengths of FL round durations.} 


\subsubsection{Summary of key results, findings that lead to development of \textit{AutoFLSat}}

From all of the experimental findings that were performed in sections \ref{subsec:accuracy}, \ref{sec:roundurr}, and \ref{subsec:idletime}, we gained several lessons crucial in developing the final algorithm protocols of \textit{AutoFLSat} in its current form.

There is the need for a \textbf{balanced scheduling optimizer}, to prevent too aggressive reductions in model training time before each round aggregation. This encouraged the development of a synchronized FL algorithm, to ensure a minimum number of epochs would be completed each round on every satellite.

\textbf{Increasing access points}, whether through more ground stations or through ISL links, \textbf{inevitably reduces the FL round duration time}. This prompted the use of satellite constellations that had at least 10 satellites per cluster, to enable the use of Intra SL links within each cluster.

The \textbf{benefit of adding more ground stations} in a network seems to have a \textbf{plateauing effect} after a certain degree. This prompted the push for an autonomous FL algorithm design, without the need for a central parameter server.

\subsection{Performance of \textit{AutoFLSat}}
\label{sec:autoflsatresults}

From all of the experimental findings that were performed in \cref{sec:results_param_exploration}, we were able to develop \textit{AutoFLSat} and compare the algorithm's performance against the leading alternatives such as \textit{FedSat}, \textit{FedSpace}, \textit{FedHAP}, and \textit{FedLEO}. Compared to other algorithms, \textit{AutoFLSat} is built to ensure proper scalability can be achieved with larger constellations, without the need for additional infrastructure such as HAPs or ground stations. 

We show comparisons of performance in \cref{tab:autoFLSat_compares}, and we find that the autonomous aggregation method provides a 12.5\%-37.5\% reduction in aggregation time compared to other leading alternatives. This is while maintaining a performance guarantee of at least 80\% accuracy on the FEMNIST and CIFAR-10 datasets, while other methodologies such as \textit{FedSpace} struggle to reach 40\% with triple the time. 

\textbf{Further explorations into the effects of clusters and epochs per round on \textit{AutoFLSat}'s performance metrics are discussed in \cref{sec:clustertests}.}

To simulate similar training scenarios with actual satellite data, the \textit{AutoFLSat} algorithm is trained on EUROSAT, which consists of image patch data taken from the Sentinel-2 spacecraft. In just 70-80 rounds of training, all satellite cluster scenarios are able to each close to convergence with results over 80\% accuracy, with total training times ranging between 6-14 hours depending on the number of clusters available on the satellite as shown in \cref{tab:eurosatautoflsat} in the appendix. This speed in model aggregation is unable to be reached by any of the models outlined in \cref{tab:autoflConfigs}, which puts \textit{AutoFLSat} at a promising place for further improvement and exploration. One thing to note however is the trend of specific cluster models suffering in relation to the global model is still present, with the high data heterogeneity present across clients. Further scrutiny will need to be put on methods to mitigate this divergence across clusters, and whether or not it it is worth having localized cluster models in addition to a larger global model to personalize the inference step of each satellite.

\setlength{\tabcolsep}{30pt}

\begin{table}[htbp]
  \setlength{\tabcolsep}{0.3\tabcolsep}
\centering
\resizebox{\linewidth}{!}{%
\begin{tabular}{c|ccc} 
\toprule
\textbf{FL Algorithm} & \textbf{Dataset} & \textbf{Accuracy} & \textbf{Training Time} \\ \midrule

AutoFLSat (4 Clusters) & FEMNIST & 83.01\% & 21.28 hours \\ 
NomaFedHAP & non-IID MNIST & 82.73\% & 24 hours \\ 
FedLEO & non-IID MNIST & 84.69\% & 36 hours \\ 
FedSat & non-IID MNIST & 85.15\% & 24 hours \\ 
FedSpace & non-IID MNIST & 52.67\% & 72 hours \\ 
\midrule
AutoFLSat (4 Clusters) & CIFAR-10 & 82.46\% & 15.6 Hours \\ 
NomaFedHAP & CIFAR-10 & 77.36\% & 24 hours \\ 
FedLEO & CIFAR-10 & 73.26\% & 36 hours \\ 
FedSat & CIFAR-10 & 81.18\% & 24 hours \\ 
FedSpace & CIFAR-10 & 39.41\% & 72 hours \\

\bottomrule
\end{tabular}
}

\vspace*{3mm}
\caption{Evaluation of \textit{AutoFLSat} with other leading FL in space algorithms, specifically trained on the FEMNIST and CIFAR-10 datasets. These two datasets have been used as a standard benchmark for performance assessment across, with metrics of final performance accuracy and total training time being contrasted against. For accuracy, \textit{AutoFLSat} is on the higher end of convergence percentages compared to the other FL in Space methods. The larger distinguishing factor however, is the faster training time found to reach the model accuracy convergence listed.}
\label{tab:autoFLSat_compares}
\end{table}

\section{Discussion}

There are several key implications that we outline for satellite constellation operators and designers that desire to perform collaborative training in orbit. 

\textbf{Designing New FL Specific Satellite Constellations.} For satellite constellations in the design phase, operators must first outline the time constraints their missions have in relation to FL training completion.
Using this constraint as a guideline, optimal FL satellite constellation architectures can be developed by modifying the ground station infrastructure or satellite hardware communication capabilities needed on board. To eliminate the problem of communication windows available completely, using a constellation design that can support an autonomous FL framework like \textit{AutoFLSat} would be ideal to provide the fastest overall model training and convergence times. 

\textbf{FL Operation with Existing Satellite Constellations.}  As discussed in \cref{subsec:accuracy}, all constellation designs should reach convergence with a high model training accuracy if provided enough communication opportunities to peer satellites and ground stations. If implementing FL onto an existing satellite constellation, the first analysis needs to consider whether or not the constellation has enough communication points to reach this model convergence within the time constraints set by the mission. If the number of satellites are to be increased within the constellation, a priority should be set on increasing the number of satellites within the same cluster rather than adding an additional orbital plane. This was discussed in \cref{sec:roundurr}, where the addition of more clusters did not correlate with shorter training durations. 

\textbf{Trade-off Computation-Communication.} The compute capabilities of an FL satellite in a LEO constellation should be chosen with careful consideration of the communication timelines in orbit and the available power budget. For example, if the computation speed significantly exceeds the communication bandwidth (and communication opportunities), then it is likely the satellite will be idle for the majority of the orbit period, since the satellite cannot perform excessive epochs in a given round without damaging the overall FL convergence.
\section{Related Work}

Finally, we discuss three domains of prior related works within the FL in space domain. This is then followed by the respective limitations of these methods with respect to real-world FL in space deployment.

\textbf{FL Algorithm Adaptations.}  First, Chen et al. \cite{chen2021satellite} considered the application of FedAvg \cite{McMahan2016CommunicationEfficientLO} to LEO constellations instead of traditional centralised ML methods. 
Afterwards, Razmi et al. proposed FedSat \cite{Razmi2022FedSat}, a method which leverages periodic client availability within the constellation to form an asynchronous implementation of FedAvg. 
In a follow-up work \cite{Matthiesen2023FedSatFollowUp}, Matthiesen et al. consider using multi-hop connections within the orbit plane and predictive routing for near persistent connectivity to ground stations.
\cite{so2022fedspace} proposed FedSpace, an extension FedBuff \cite{nguyen2022federated} which considers using ground stations as parameter servers.
\textbf{Limitation:} In comparison to our space-ification framework, no generalization of adapting these FL protocols for space was provided in these FL algorithm adaptations. Different approaches were taken for every reconfiguration, without standardization for each methodology.

\textbf{FL in Space Platforms} A new scenario for FL in satellite constellations was proposed by Elmahallawy and Luo \cite{elmahallawy2022fedhap}, leveraging high-altitude platforms (HAPs) to form a hierarchical FL structure in which parameters are exchanged in two tiers, namely intra-plane satellite links (the \textit{worker} tier), and HAPs (the \textit{server} tier). \textbf{Limitation:} To fully implement FedHAP, additonal cost is needed to build out this infrastructure. Compared to FedHAP, \textit{AutoFLSat} is built to ensure proper scalability can be achieved with larger constellations, without the need for additional infrastructure such as HAPs or ground stations.

\textbf{Intra and Inter plane Links} More recent works such as FedLEO \cite{Zhai2024FedLEO} and FedMega \cite{shi2024satellite} emphasise the benefits of fully decentralised FL in which satellites can arbitrarily take on the role of FL clients, parameter servers, or both. Here, a new focus is placed on leveraging inter-plane satellite links that are brief and highly variable in nature. \textbf{Limitation: } Inter-plane links open a range of possibilities for FL in space due to added flexibility in the constellation design, but are subject to data rate constraints due to their short intervals. These last 2 works do not employ realistic hardware constraints and only consider the optimal scenario assumptions to employ inter-plane link useage.

\section{Conclusions}

This work provides a holistic overview of the key considerations needed for FL applications in space. First, we introduce theoretical space-ification framework with modular optimizations that can improve the performance of any algorithm with space-ification. We evaluate these on FLySTacK, 
showcasing results of an extensive satellite configuration space exploration.
Finally, this leads into the development and introduction of \textit{AutoFLSat}, a novel autonomous FL algorithm for space was developed that took into account the discovered benefits of ISL links, balancing scheduling optimization and accuracy, and focus on FL round duration rather than idle time. 


\bibliography{references}
\bibliographystyle{mlsys2024}

\newpage

\appendix
\appendix
\section{Appendix}
\subsection{Space-ification: an orbital suite of FL algorithms in space}\label{appen:spacification_fedprox_fedbuff}

\subsubsection{Space-ification of the \textit{FedProx} Algorithm: Allowing for Partial Updates}

With satellites in the same constellation revisiting ground stations at different times, it is not possible to reach perfect synchronicity for client parameter aggregation. To address this, the \textit{FedProx} algorithm was selected to allow for the incorporation of partial client updates. By effectively utilizing this partial update scheme, a significant amount of idling time would be eliminated from each client. Despite being a synchronous FL method, \textit{FedProx} provides more flexibility for clients with less optimal orbital schedules to join each round. Compared to standard FL algorithms like \textit{FedAvg}, each client is not forced to complete the same number of training epochs each round.
\vspace{0.3mm}
\begin{algorithm}[!ht]
\caption{\textit{\highlight{FedProxSat}}. \textit{The base algorithm in black is referenced from the original \textit{FedProx} paper \cite{li_federated_2020}, with modifications for satellite specific operations in \highlight{blue}. \textit{FedProx} specific changes are in \highlightgreen{purple}.}\newline
The $K$ satellites (clients) are indexed by $k$, $C$ for the maximum number of clients averaged each round, $B$ is the local minibatch size, $E$ is the number of local epochs, $\mu$ is the learning rate, $P_k$ set of indexes of data points in client $k$.}\label{alg:fedProxSat}
\begin{algorithmic}
\algrenewcommand\algorithmicrequire{\textbf{Main Server executes:}}
\algrenewcommand\algorithmicensure{\textbf{ClientUpdate($k$,$w$):}}
\Require 
    \State initialize $w_0$ (global model weights)
\For{each round $t = 0,1,2,...$} 
    \State{$m \leftarrow$ \highlight{min}$(C, K)$} \Comment{\textit{\highlight{Satellites typically don't have enough clients}}}
    \State{$S_t \leftarrow $\highlight{(first $m$ clients that contacts a ground station, and not currently training)}}
    \State{\textbf{Send} $w_t$ to each $k$ in $S_t$}
    \For{each client $k \in S_t$ \textbf{in parallel}}
        \State{$w^k_t$ =  ClientUpdate $(k,w_t)$}
        \State{\highlight{\textbf{Wait} for contact with $k$, client continues to train until it reaches ground station}}
        \State{$w^k_{t+1} \leftarrow$ $w^k_t$}
    \EndFor
    \State{$m_t \leftarrow \Sigma_{k \in S_t} n_k$}
    \State{$w_{t+1} \leftarrow \Sigma_{k \in S_t} \frac{n_k}{m_t} w^k_{t+1}$}
\EndFor
\newline

\Ensure \Comment{\textit{Run on client/satellite k}}
\State {\highlight{\textbf{Receive}  $w$ from ground station}}
\State{$\beta \leftarrow$ (Split $P_k$ into batches of size $B$)}
\While{\highlightgreen{no access to ground station, in local epoch}}
\For{ batch $b \in \beta$ }
\State{$w_t = \mu \nabla l(w;b)$}
\State{$w \leftarrow w_t + \frac{\mu}{2} || w_t - w ||^2$} \Comment{\textit{\highlightgreen{Proximal Term, limits impact of local updates}}}
\EndFor
\EndWhile
\State{\highlight{return $w$ to nearest ground station}}
\end{algorithmic}
\end{algorithm}

As mentioned in \cref{sec:spaceify}, similar modifications to the original \textit{FedAvg} algorithm were also made in \textit{FedProx}. Client selection is conducted in the same manner, and no scheduling is considered in this process. In \cref{alg:fedProxSat}, the protocol for operations are outlined. The main difference between \textit{FedAvg} and \textit{FedProx} is in the client function, where now the client continues to train for as many epochs as possible, rather than a predefined number from the server. The updates from the trained local client model are limited by a proximal term comparing the deviation of the update to the original global parameters that were sent to client at the start of the round.

\subsubsection{Space-ification of the \textit{FedBuff} Algorithm: Asynchronous Aggregation}

\textit{FedBuff} provides a different lens of implementation, due to it's asynchronous methodology. Similar to \textit{FedProx}, each client is allowed to train for more epochs, while providing weights that have only completed partial updates.

\begin{algorithm} [!ht]
\caption{\textit{\highlight{FedBuffSat}}. \textit{The base algorithm in black is referenced from the original \textit{FedBuff} paper \cite{nguyen2022federated}, with modifications for satellite specific operations in \highlight{blue}. \textit{FedBuff} specific changes are in \highlightgreen{purple}.}
\newline
Buffer size $D$, $K$ satellites (clients) indexed by $k$, $C$ for the maximum number of clients averaged each round, $B$ is the local minibatch size, $E$ is number of local epochs, $\mu$ is the learning rate, $P_k$ set of indexes of data points in client $k$.}\label{alg:fedBuffSat}
\begin{algorithmic}
\algrenewcommand\algorithmicrequire{\textbf{Main Server executes:}}
\algrenewcommand\algorithmicensure{\textbf{ClientUpdate($k$,$w$):}}
\Require 
    \State initialize $w_0$ (global model weights)
\While{not converged, each round $t = 0,1,2,...$} 
    \State{\highlightgreen{Set buffer size} $D \leftarrow$ \highlight{min}$(C, K)$} \Comment{\textit{\highlightgreen{Buffered Update, for secure aggregation}}}
    \State{$c \leftarrow $(\highlight{clients contact a ground station})} \Comment{\textit{\highlightgreen{Async}}}
    \State{Run \textbf{ClientUpdate} on clients $c$} \Comment{\textit{\highlightgreen{Async}}}
    \If {client contacts to update}
        \State{$w^k_t$ =  ClientUpdate $(k,w_t)$}
        \State{$w^k_{t+1} \leftarrow$ $w^k_t$}
        \State{$d = d+1$}
        \Comment{\textit{\highlightgreen{Updating current buffer size}}}
        
        \If {$d == D$}
            \State{$m_t \leftarrow \Sigma_{k \in S_t} n_k$}
            \State{$w_{t+1} \leftarrow \Sigma_{k \in S_t} \frac{n_k}{m_t} w^k_{t+1}$}
            \State{Completed round, reset buffer}
        \EndIf
    \EndIf
\EndWhile 
\newline

\Ensure \Comment{\textit{Run on client/satellite k}}
\State {\highlight{\textbf{Receive}  $w$ from ground station}}
\State{$\beta \leftarrow$ (Split $P_k$ into batches of size $B$)}
\While{\highlightgreen{no access to ground station, in local epoch}}
\For{ batch $b \in \beta$ }
\State{$w_t = \mu \nabla l(w;b)$}
\State{$w \leftarrow w_t + \frac{\mu}{2} || w_t - w ||^2$} \Comment{\highlightgreen{\textit{Similar to \textit{FedProx}, Proximal Term, limits impact of local updates}}}
\EndFor
\EndWhile
\State{\highlight{return $w$ to nearest ground station}}
\end{algorithmic}
\end{algorithm}

However, the aggregation technique works on an asynchronous buffer system, resulting in each client not synchronizing to the same global model. As each client reaches the server at different time points in the round, satellites are allowed to continue training until they reach back out to a ground station. This reduces almost all computational idle time on the satellites. Depending on how long it takes for these satellites to reach a ground station, this may lead to some clients having outdated global models. The results from outdated models are only added if they are under the bounded staleness metric set by the global server. The protocol is outlined in \cref{alg:fedBuffSat}.

\section{FL Augmentation Algorithms}\label{appen:fl_schedule_alg}\label{appen:fl_intrasl_alg}
The FL Schedule algorithm is outlined in Algorithm \ref{alg:fedAvg2Sat}.
\begin{algorithm}[!ht]
\caption{\textit{\highlight{FLSchedule}}.  \textit{Once again, the base algorithm in black is referenced from the original \textit{FedAvg} paper \cite{mcmahan_communication-efficient_2016}, with modifications for satellite specific operations in \highlight{blue}.} \newline The $K$ satellites (clients) are indexed by $k$, the $G$ ground stations (servers) are indexed by $g$, $C$ for the maximum number of clients averaged each round, and $L$ indicates orbital data/geographic location of the respective satellite/ground station.}\label{alg:fedAvg2Sat}
\begin{algorithmic}
\algrenewcommand\algorithmicrequire{\textbf{Main server executes:}}
\algrenewcommand\algorithmicensure{\textbf{ClientUpdate($k$,$w$):}}
\algnewcommand\Scheduler{\item[\textbf{Scheduler}$(L_k, L_g, C, t)$]}
\Require 
    \State{\textit{Use appropriate initialization protocols from specific FL alg}}
\For{each round $t = 0,1,2,...$} 
    \State{$S_t \leftarrow$ Scheduler$(L_k,L_g,C,t)$ }\Comment{\textit{\highlight{Select clients using scheduler}}}
    \State{\textbf{Send} $w_t$ to each $k$ in $S_t$}
    \For{each client $k \in S_t$ }
    \State{\textit{Use appropriate aggregation protocols from specific FL alg}}
    \EndFor

\EndFor
\newline
\highlight{
\Scheduler
\State{Initialize $S_t$}
\Comment{\textit{Client list}}
\State{$T \leftarrow $ Start time $t$}
\Comment{\textit{Set start time of orbit \textbf{propagation}}}
\While{All clients not selected}
\State{\textbf{Propagate} each satellite client $k$'s orbital trajectories}
\If{$k$ can contact any ground station $g$}
    \State{\textbf{Calculate} $k$ to $g$ access times}
    \If{second time $k$ has contacted any $g$, and k is not in $S_t$}
        \State{$S_t \leftarrow$ append $k$} \Comment{\textbf{Schedule clients with \textbf{fastest} contact and return)}}
        \If{length of $S_t == C$}
            \State{Return $S_t$}
        \EndIf
    \EndIf
\EndIf
\EndWhile}
\newline

\Ensure 
\Comment{\textit{Use appropriate \textbf{ClientUpdate} function from specific FL alg.}}

\end{algorithmic}
\end{algorithm}

The FL IntraSL algorithm is outlined in Algorithm \ref{alg:fedAvg3Sat}.

\begin{algorithm}[!ht]
\caption{\textit{\highlight{FLIntraSL}}. \textit{Once again, the base algorithm in black is referenced from the original \textit{FedAvg} paper \cite{mcmahan_communication-efficient_2016}, with modifications for satellite specific operations in \highlight{blue}.} The $K$ satellites (clients) are indexed by $k$, the $G$ ground stations (servers) are indexed by $g$, $C$ for the maximum number of clients averaged each round, and $L$ indicates orbital data/geographic location of the respective satellite/ground station.}\label{alg:fedAvg3Sat}
\begin{algorithmic}
\algrenewcommand\algorithmicrequire{\textbf{Main server executes:}}
\algrenewcommand\algorithmicensure{\textbf{ClientUpdate($k$,$w$):}}
\algnewcommand\Scheduler{\item[\textbf{IntraSL}$(L_k, L_g, C, t)$]}
\Require 
    \State{\textit{Use appropriate initialization protocols from specific FL alg}}
\For{each round $t = 0,1,2,...$} 
    \State{$S_t \leftarrow$ \textbf{IntraSL$(L_k,L_g,C,t)$} }\Comment{\textit{\highlight{Select clients using IntraSL Scheduler}}}
    \State{\textbf{Send} $w_t$ to each $k$ in $S_t$}
    \For{each client $k \in S_t$ }
        \State{\textit{Use appropriate aggregation protocols from specific FL alg}}
    \EndFor
\EndFor
\newline

\highlight{
\Scheduler
\State{Initialize $S_t$}
\Comment{\textit{Client list}}
\State{Initialize for each cluster $S_{passed}$}
\Comment{\textit{Each cluster, track previously seen satellites}}
\State{$T \leftarrow $ Start time $t$}
\Comment{\textit{Set start time of orbit \textbf{propagation}}}
\While{all clients not selected}
\State{\textbf{Propagate} each satellite client $k$'s orbital trajectories}
\newline
\If{$k$ can contact any ground station $g$}
    \State{\textbf{Calculate} $k$ to $g$ access times}    
    \State{\textbf{Add} $k$ to their cluster's $S_{passed}$}
    \newline
    \If{second time $k$ has contacted any $g$, and $k$ is not in $S_t$}
        \State{$S_t \leftarrow$ append $k$} \Comment{\textbf{Schedule} clients with \textbf{fastest} contact and return}
        \If{length of $S_t = C$}
            \State{Return $S_t$}
        \EndIf
    \EndIf
    \newline
    \If{any other same cluster satellite $k_c$ listed in $S_{passed}$ and $k_c$ not in $S_t$}
        \State{$S_t \leftarrow$ append $k$} \Comment{\textit{Use \textbf{IntraSL} comms to send back model parameters}}
        \If{length of $S_t == C$}
            \State{Return $S_t$}
        \EndIf   
    \EndIf
\EndIf
\EndWhile
}
\newline

\Ensure 
\Comment{\textit{Use appropriate \textbf{ClientUpdate} function from specific FL alg.}}

\end{algorithmic}
\end{algorithm}

\section{FLyCubes - An FL in Space Platform Implementation}
\label{sec:FLyCubes}
Much of the advancement in space accessibility in the 21st century can be attributed to CubeSats, standardized low-cost and lightweight nanosatellites \cite{Heidt2000CubeSatAN}. To explore the hardware constraints affecting FL applications in space, we focused our efforts on developing a 1st-generation CubeSat prototype with FL capabilities. We refer to our design as the \textbf{FLyCube} (\textbf{F}ederated-\textbf{L}earning Read\textbf{y} \textbf{Cube}Sat). 

\subsection{FLyCube Design} 
\begin{figure}[!htb]
    \centering
    \includegraphics[width=1\linewidth,trim={0.2cm 0 0.2cm 0},clip]{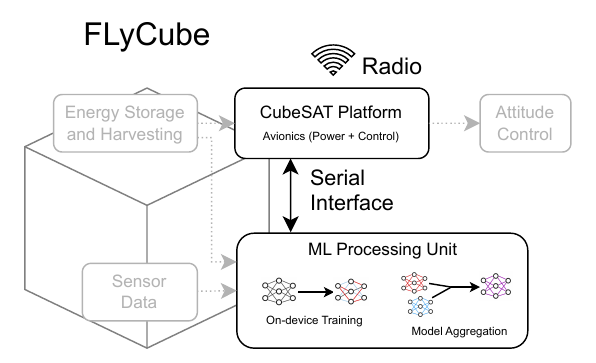}
    \caption[FLyCube system high-level overview.]{FLyCube system overview. The CubeSat platform and ML processing unit used would inform the design of an energy management system, actuators and onboard sensors, along with the downstream application in question.}
    \label{fig:flycube_block_diagram}
\end{figure}

\textbf{Design Approach.} FLyCube's design is focused on rapid prototyping and sub-component interchangeability. To this end, the system comprises a separated CubeSat platform and ML processing unit (Figure \ref{fig:flycube_block_diagram}) using open-source and commercial off-the-shelf (COTS) components. Note CubeSat platform here refers to the primary avionics board which would control essential satellite functions, namely power management, driving peripheral sensors, attitude control, and communications (only the communication functionality will be utilized for FL experiments). The ML processing unit should perform FL tasks: on-device inference and training, and, in the case of fully decentralized FL, model aggregation.

\textbf{CubeSat Platform.} For FLyCube's CubeSat platform, we use the open-source PyCubed framework \cite{Holliday2019PyCubed}, noting that it facilitates rapid prototyping and development. In particular, the PyCubed system is programmable using CircuitPython, with open-source access to PCB fabrication and assembly (PCBA) files and various flight software examples.

\textbf{ML Processing Unit} The FLyCube employs an additional processing element (a co-processor), configured as a peripheral, which performs ML tasks independently of the CubeSat platform. Our implementation considers a Raspberry Pi Zero 2W for the ML processing unit, but this could be changed for alternative COTS solutions such as the NVidia Jetson series due to a standardized USB interface to the CubeSat platform.

\textbf{Parameter Communications} For FL tasks, the FLyCube must be capable of directional parameter communication. Internal parameter communication, i.e., between the ML processing unit and CubeSat platform, are handled by the USB serial connection, where external parameter communication (i.e., between FLyCube systems or to ground stations) is performed via long-range (LoRa) radio. We employ a 433MHz UHF-band RFM98PW radio module for compatibility with the PyCubed framework \cite{Holliday2019PyCubed}.

\textbf{Power} The ML processing unit (Raspberry Pi Zero 2W) receives USB power from a 12.5W power supply. To independently monitor the power consumption of the FLyCube system under varying workloads, we employ Adafruit INA260 current sensors at the input port, where current and voltage readings are logged to a separate machine using a Raspberry Pi Pico microcontroller. Our analysis of FLyCube's power usage considers the consumption during various FL workloads, as well as the added orbital average power (OAP) introduced by FL tasks (Table \ref{tab:power_modes}). Our analysis of FLyCube’s power usage considers the consumption during various FL workloads, as well as the added orbital average power (OAP) introduced by FL tasks (Table \ref{tab:power_modes}).

\begin{table*}[!htb]
    \centering
    \footnotesize
\resizebox{1\linewidth}{!}{%
    \begin{tabular}{l r c c}
    \toprule
        \textbf{Power Mode} & \textbf{Consumption (mW)} & \textbf{Duty Cycle} & \textbf{OAP (mW)} \\

    \cmidrule(lr){1-4}
    
        \multicolumn{4}{l}{\textbf{CubeSat Framework (PyCubed)}} \\
        \cmidrule(lr){1-2}
        Idle & 207\tiny{$\pm$8.1} \\
        Radio TX\footnotemark & 960\tiny{$\pm$22.3} \\
    
    \cmidrule(lr){1-2}
    
        \multicolumn{4}{l}{\textbf{ML Processing Unit (Raspberry Pi Zero 2W)}} \\
        \cmidrule(lr){1-2}
        Off & 191\tiny{$\pm$5.4} \\
        Idle & 653\tiny{$\pm$18.6} \\
        Stress (Training) & 2178\tiny{$\pm$33.7} \\
    
    \cmidrule(lr){1-4}

    \multicolumn{4}{l}{\textbf{FLyCube (PyCubed + Raspberry Pi Zero 2W)}} \\
    \cmidrule(lr){1-4}
        Low Power (Idle) & 760\tiny{$\pm$6.8} & 0 & 0 \\
        Radio TX & 1613\tiny{$\pm$20.5} & 0 & 0 \\
        Training & 2178\tiny{$\pm$20.9} & 0.8 & 1742 \\
        High Power (Training + Radio TX) & 3138\tiny{$\pm$28.0} & 0.2 & 628 \\
    \midrule
        \textbf{Total} & & \multicolumn{2}{r}{\textbf{2370 mW}} \\
    \bottomrule
    \end{tabular}
}
    \caption[FLyCube power modes.]{FLyCube power modes. Idle power consumption is approximately 0.76W, with a peak of 3.2W during simultaneous training and radio transmission. Assuming a constellation of five FLyCubes (20\% of time spent transmitting models), FLyCubes' decentralized learning functionality adds approximately 2.4W to the consumed orbital average power (OAP).}
    \label{tab:power_modes}
\end{table*}

\subsection{Flight Software}
PyCubed offers programmability in CircuitPython, a low-level Python interpreter that parses Python into hardware-level instructions for embedded processors such as the PyCubed SAMD51. Accessing low-level hardware interfaces such as I2C, SPI, and USB is possible with simple Python syntax, allowing rapid testing of various hardware configurations. CircuitPython additionally supports \texttt{asyncio} for asynchronous I/O task handling. We employ the open-source PyCubed V0.5 firmware leveraging CircuitPython v8.2.0 \cite{Holliday2019PyCubed}.

\subsection{Decentralized Learning Pipelines}
The ML processing unit handles FL tasks (on-device training and model aggregation) onboard the FLyCube system, while also responding asynchronously to serial commands from the PyCubed framework. Our decentralized learning pipelines are implemented in Python for rapid prototyping. Our Python implementation, paired with the standardized USB serial interface between the ML processing unit and CubeSat platform, makes the ML processing unit interchangeable with any COTS solution capable of running Python with minimal modification of source code.

\textbf{On-device Training.} The FLyCube system uses PyTorch for on-device training and evaluation.

\begin{figure*}[!htb]
    \centering
    \includegraphics[width=0.8\linewidth]{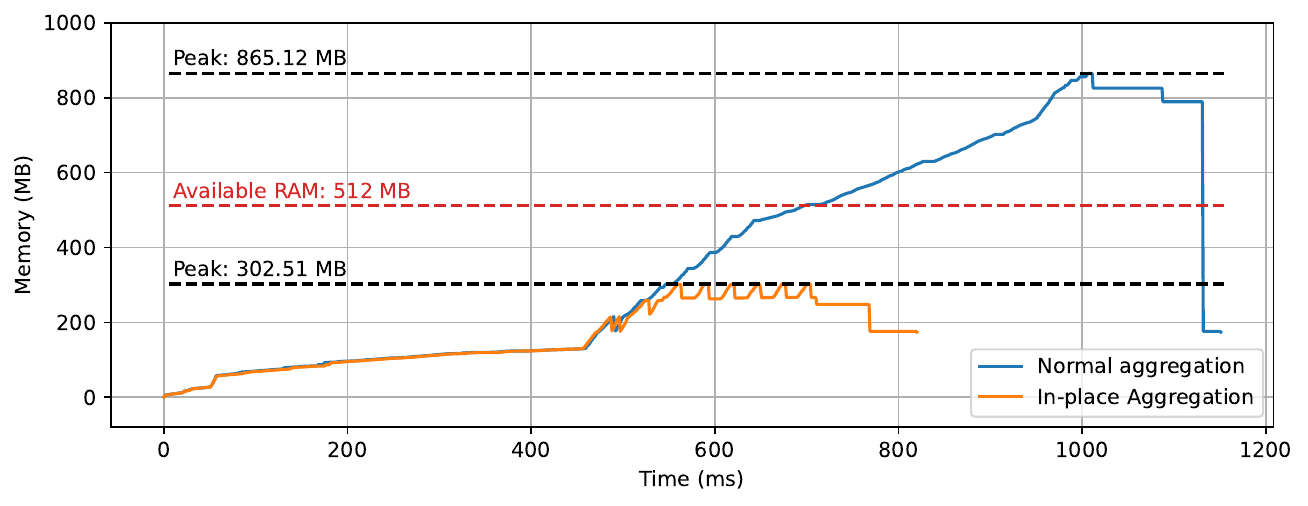}
    \caption[Aggregating eight ResNet18 models onboard FLyCubes using conventional and in-place model aggregation methods.]{Aggregating eight ResNet18 models onboard FLyCubes using conventional and in-place model aggregation methods. By accumulating the aggregated parameters in fixed memory, in-place aggregation is faster and more memory efficient than conventional model aggregation, avoiding the need to use swap memory from the comparatively slow SD card when available system RAM is exceeded.}
    \label{fig:in_place_aggregation_results}
\end{figure*}

\textbf{Model Aggregation.} FLyCubes leverage the in-place aggregation method from the Flower FL framework \cite{beutel_flower_2020}. In-place aggregation uses a fixed amount of memory to accumulate the aggregated model from a number of updated client models, resulting in reduced memory usage and faster aggregation. Without in-place aggregation, we risk exceeding the 512MB of SDRAM on the Raspberry Pi Zero 2W at aggregation time, resulting in a severe penalty for using the slow SD card as swap memory, illustrated in Figure \ref{fig:in_place_aggregation_results}.

\subsection{Characterizing the FLyCube System}
The FLyCube system is characterized by a maximum effective radio CubeSat-to-CubeSat data rate of roughly 1.6 KB/s and less than 3.5W peak power consumption during simultaneous on-device training and parameter transmission over radio. FLyCube's FL functionality adds roughly 2.4W orbital average power (OAP) consumed, assuming a constellation of five FLyCubes and a Raspberry Pi Zero 2W ML processing unit. The combination of the PyCubed and Raspberry Pi Zero 2W occupies roughly 25\% of a 1U CubeSat structure, facilitating the future inclusion of batteries and dedicated power delivery circuitry. With our decentralized learning pipelines, the FLyCube system can perform on-device training and model aggregation within the 512MB SRAM constraints of the Raspberry Pi Zero 2W.

\subsection{Federated Learning with FLyCubes}
\begin{figure}[H]
    \centering
    \includegraphics[width=1\linewidth,trim={0 2cm 0 0},clip]{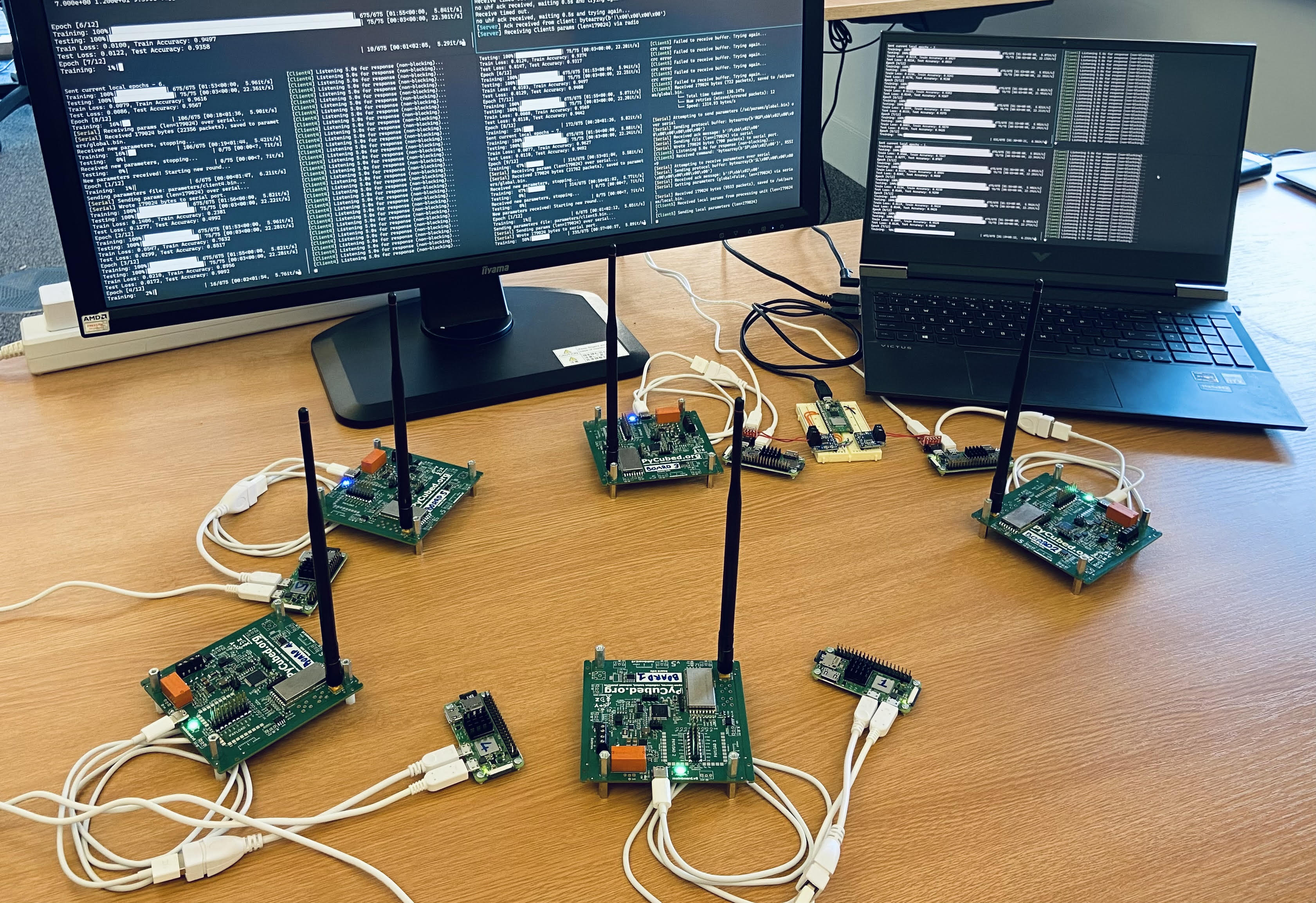}
    \caption{Federated Learning with five FLyCube prototypes.}
    \label{fig:sys_demonstration}
\end{figure}

We demonstrate practical FL on a constellation of five built FLyCube prototypes in Figure \ref{fig:sys_demonstration}.

\textbf{Experimental Setup.} Our FL experiments consider Zakerinia et al.'s Quantized Asynchronous Federated Learning (QuAFL) algorithm \cite{Zaker2023QuaFL}. QuAFL is an asynchronous implementation of FedAvg, allowing any arbitrary number of clients to be sampled asynchronously each round while also employing quantization to reduce communication overheads. Given FLyCubes low data rate of 1.6 KB/s, we consider running QuAFL on a single cluster (orbit plane) comprising five clients corresponding to the number of FLyCube prototypes built. A single subsequent client is sampled per round, mimicking the order of contact to the ground station. 

\textbf{Results.} The FLyCubes FL results on the EuroSAT classification task are shown in Table \ref{tab:fl_timelines_orbit}.

\begin{table*}[!htb]
    \centering
    \footnotesize
    
    \resizebox{\textwidth}{!}{%
    \begin{tabular}{lccccc}
        \toprule

         & Centralized & \multicolumn{4}{c}{\textbf{QuAFL}}\\
        \cmidrule(lr){3-6}
        \cmidrule(lr){2-2}
        
        \textbf{Model} & \textbf{Test Acc. (\%)} & \textbf{Precision} & \textbf{Test Acc. (\%)} & \textbf{RTC} & \textbf{WCTC} \\

        \midrule
        
        \multirow{3}{*}{LeNet5} & \multirow{3}{*}{79.3} 
        & 32-bit (full) & 69.5 & 25 & 1d 5h 22m \\
        \cmidrule(lr){3-6}
        &  & 10-bit & \textbf{69.4} & \textbf{25} & \textbf{9h 10m} \\
        \cmidrule(lr){3-6}
        &  & 8-bit & 68.0 & 39 & 11h 27m \\

        \midrule
    
        \multirow{3}{*}{MobileNetV2} & \multirow{3}{*}{92.7} 
        & 32-bit (full) & 75.4 & 15 & 4d 21h 10m \\
        \cmidrule(lr){3-6}
        &  & 10-bit & \textbf{75.0} & \textbf{15} & \textbf{4d 14h 0m} \\
        \cmidrule(lr){3-6}
        &  & 8-bit & 73.7 & 40 & 12d 5h 20m \\

        \midrule
        
        \multirow{3}{*}{ResNet18} & \multirow{3}{*}{87.1} 
        & 32-bit (full) & 73.2 & 9 & 14d 13h 25m \\
        \cmidrule(lr){3-6}
        &  & 10-bit & \textbf{72.7} & \textbf{9} & \textbf{4d 13h 11m} \\
        \cmidrule(lr){3-6}
        &  & 8-bit & 71.2 & 22 & 8d 21h 32m \\        
        \bottomrule
    \end{tabular}
    }
    
    \vspace{3mm}
     \caption{FLyCubes FL results on EuroSAT dataset. We show the centralized EuroSAT test accuracy of each model, number of rounds to converge (RTC), and wall clock time to converge (WCTC). Quantizing models to 10-bit precision at communication time offers the optimal balance between raw transmission time and convergence rate on EuroSAT data while closely matching full-precision performance.}
     \label{tab:fl_timelines_orbit}
\end{table*}

\subsection{Discussion}\label{sec:flycubes_discussion}
Table \ref{tab:fl_timelines_orbit} shows that the FLyCubes constellation can train both ResNet18 and MobileNetV2 on decentralized EuroSAT data within five calendar days. This translates to less than 1\% of nanosatellite mission lifetimes, typically ranging from two to five years \cite{DISCOS}. If we consider the worst cases of wall clock time to convergence in Table \ref{tab:fl_timelines_orbit}, emulating cases where the optimal bit-precision may not be known at mission-time, convergence still occurs over the course of roughly two weeks; 3\% of mission lifetime. From this, we can conclude that fully decentralized learning via intra-plane satellite links, i.e., within a single orbit plane, is practically feasible with our developed FLyCube constellation. However, the applicability of FLyCubes in larger constellations with inter-plane links is dependent on a number of key focus areas, discussed next.

\textbf{The Data Rate Bottleneck.} Despite selecting low-power ML processing unit (Raspberry Pi Zero 2W) for reduced material costs and power consumption, wall clock time to convergence is still bottlenecked by model transmissions due to FLyCube's low radio data rate. Note the developed constellation of five FLyCubes (configured in a single cluster) falls on the smallest end of the spectrum of potential constellation sizes; if the constellation size is increased, propagation times may grow to exceed on-device training times by orders of magnitude. While quantization reduced communication times by $\approx3\times$, it also degrades the model convergence rate. In contrast, even modest improvements to the radio data rate would dramatically increase the scope of possible decentralized learning tasks in orbit while also having no direct effect on model convergence.

\begin{figure*}[htbp]
\begin{minipage}{\textwidth}
\centering
\begin{subfigure}{.35\textwidth}
    \centering
    \includegraphics[width=0.95\linewidth]{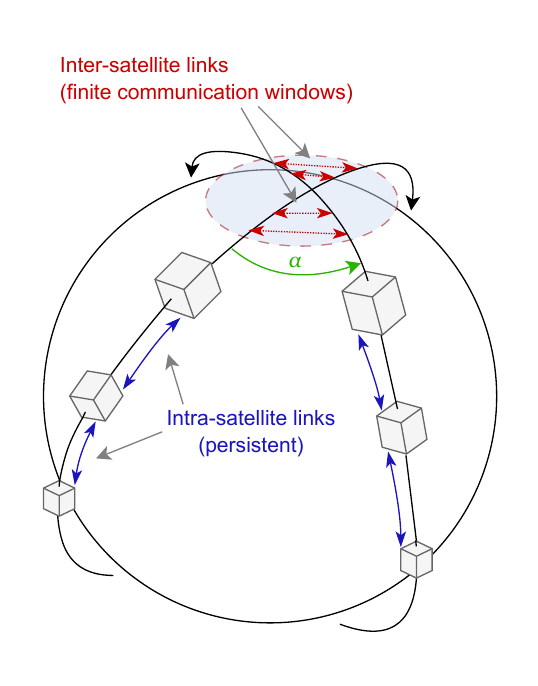}
    \caption{Relative Plane Angle $\alpha$}
    \label{fig:alpha_illustration}
\end{subfigure}%
\begin{subfigure}{.65\textwidth}
    \centering
    \includegraphics[width=0.86\linewidth]{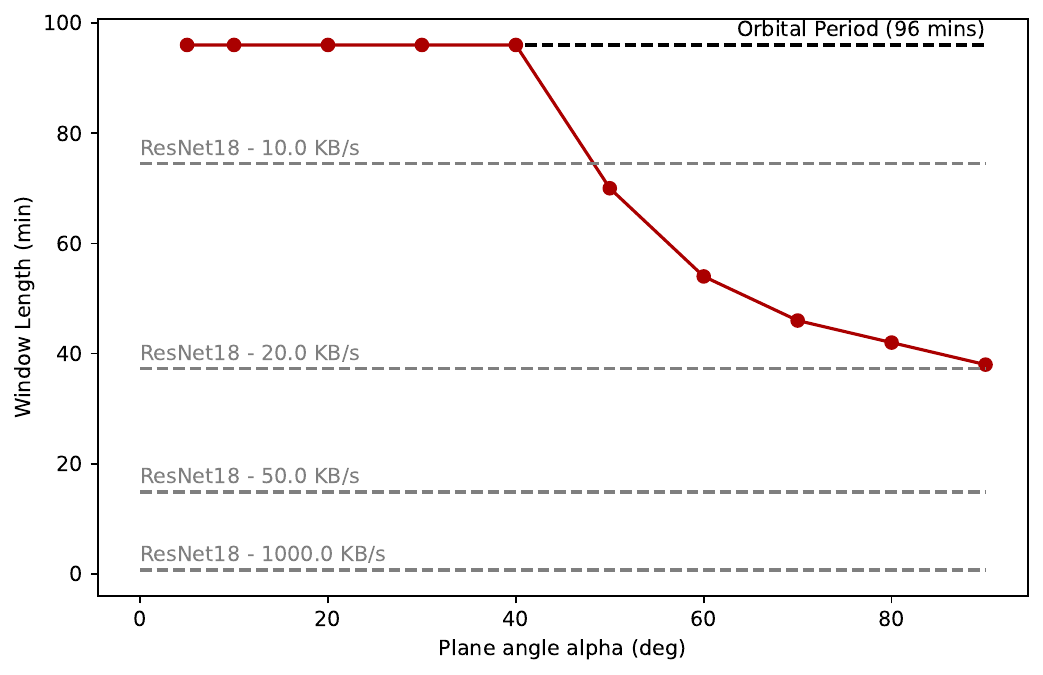}
    \caption{Effect of $\alpha$ on inter-plane communication window length}
    \label{fig:alpha_plot}
\end{subfigure}%
\vspace{1\baselineskip}
\caption{Effect of relative plane angle $\alpha$ on inter-plane communication window length, assuming an orbital altitude of 400km and satellites following the same relative positions on their respective orbit paths. }
\label{fig:alpha_angle}
\end{minipage}
\end{figure*}

\textbf{Inter-plane Communications.} Data rate plays an important role for inter-plane (i.e., between cluster) communications, as it must be sufficiently high to allow model transmission within a given intra-plane communication window. The length of a communication window between two orbit planes depends on the relative positions of satellites along their orbit paths, as well as the relative plane angle $\alpha$ at the point of intersection. This is illustrated in Figure \ref{fig:alpha_illustration}, where the allowable plane angles $\alpha$ under varying data rate is shown in Figure \ref{fig:alpha_plot}.

Figure \ref{fig:alpha_plot} shows that satellites from neighboring clusters maintain LOS for the duration of the orbit period given relative plane angle $\alpha<40^\circ$, allowing near-persistent communication in constellations with 4 or more clusters in a Walker-star configuration. A minimum data rate of 20 KB/s is required to facilitate unidirectional inter-plane transmission of ResNet18 parameters at full precision regardless of $\alpha$.

\textbf{Re-assessing the Satellite GPU Debate.} In situations where communication is the bottleneck, a natural response would be to shift the workload from communication to computation. This calls for a reconsideration of the `faster is better' narrative often used to motivate the integration of GPUs onboard small satellites. We now reconsider this narrative in the context of our developed FLyCube system. Given a higher data rate, slow CPU training on the FLyCube ML processing unit (Raspberry Pi Zero 2W) would likely be the limiting factor governing wall clock time to convergence. In this scenario, a low-power device with a GPU, such as the NVidia Jetson Nano, might be a more appropriate choice for the ML processing unit. However, we reiterate that as the cluster size is increased, the model propagation times may grow to exceed on-device training times by orders of magnitude. In such cases, integrating GPUs onboard small satellites may cause unacceptably long idle times, despite the corresponding increase in design complexity, unit cost (which scales with constellation size) and power consumption. We recommend closely assessing model propagation timelines when designing constellations for decentralized learning in orbit, and using them to inform the choice of both ML \textit{and} communications hardware.

\clearpage

\section{Additional Information}

\begin{figure}[htb]
    \centering
    \includegraphics[width=0.95\linewidth]{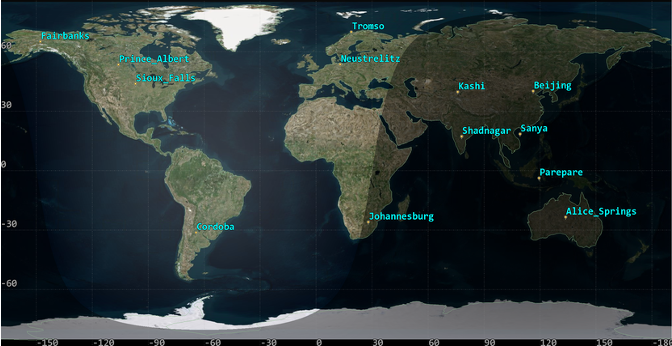}
    \vspace*{3mm}
    \caption{Ground station facilities used for parameter sweep simulations, with 13 communication locations inspired from the International Ground Station (IGS) network \cite{wulder_global_2016-1}. The different locations are set to be in: Sioux Falls (US), Sanya (China), Johannesburg (South Africa), Cordoba (Argentina), Tromso (Norway), Kashi (China), Beijing (China), Neustrelitz (Germany), Parepare (Indonesia), Alice Springs (Australia), Fairbanks (US), Prince Albert (Canada), and Shadnagar (India).}
    \label{fig:groundstations}
\end{figure}

\subsection{Experimental Setup}
\label{sec:tabofexperiments}

Every run had 5 repeated trials, which were averaged within the results. Also, the simulation of the satellite orbital trajectories were set to scenarios of three months of time starting from the 14th of April 2024, to July 13th 2024. All FL algorithms were limited in the number of training rounds based on the length of the satellite scenario simulation, or when they reached 500 rounds of training. Whichever condition that was met first was set as the stopping point of the satellites.

\setlength{\tabcolsep}{30pt}

\begin{table*}[htbp]
    \renewcommand{\arraystretch}{1.0}
    \centering
    \footnotesize
  \setlength{\tabcolsep}{0.3\tabcolsep}
\centering
\resizebox{0.92\linewidth}{!}{%
\begin{tabular}{cccccc}
\toprule
\textbf{FL Alg} & \textbf{Satellite Extension} & \textbf{Clusters} & \textbf{Satellites per Cluster} & \textbf{Ground Stations} \\ \midrule
FedAvg & Base Satellite  & $1,2,5,10$ & $1,2,5,10$ &  $1,2,3,5,10,13$ \\
FedAvg & Schedule  & $1,2,5,10$ & $1,2,5,10$ &  $1,2,3,5,10,13$ \\
FedAvg & Intra Sat Links  & $1,2,5,10$ & $1,2,5,10$ &  $1,2,3,5,10,13$ \\
\midrule
FedProx & Base Satellite  & $1,2,5,10$ & $1,2,5,10$ &  $1,2,3,5,10,13$ \\
FedProx & Schedule  & $1,2,5,10$ & $1,2,5,10$ &  $1,2,3,5,10,13$ \\
FedProx & Schedule V2  & $1,2,5,10$ & $1,2,5,10$ &  $1,2,3,5,10,13$ \\
FedProx & Intra Sat Links  & $1,2,5,10$ & $1,2,5,10$ &  $1,2,3,5,10,13$ \\
\midrule
FedBuff & Base Satellite  & $1,2,5,10$ & $1,2,5,10$ &  $1,2,3,5,10,13$ \\

\bottomrule
\end{tabular}
}
\caption{Parameter configuration settings set for testing of various satellite constellations and FL algorithms. The main sweeps were performed over the \# of clusters in a constellation, the \# of satellites per cluster, and \# of ground stations available for communication. These were applied across "space-ified" algorithms \textit{FedAvg}, \textit{FedProx}, and \textit{FedBuff} with several augmentations. All were tested on the FEMNIST Dataset.}
\label{tab:clflConfigs}
\end{table*}

\vspace{3mm}

\setlength{\tabcolsep}{30pt}

\begin{table*}[htp]

  \setlength{\tabcolsep}{0.3\tabcolsep}
\centering
\resizebox{0.8\linewidth}{!}{%
\begin{tabular}{cccccc}
\toprule
\textbf{FL Alg} & \textbf{Communication Method} & \textbf{Dataset} &  \textbf{Epochs} &\textbf{Clusters} \\ \midrule
AutoFLSat & Intra/Inter Sat Links & FEMNIST & $20,50,100,250$ & $2,3,4$ \\

AutoFLSat & Intra/Inter Sat Links & CIFAR-10 & $1,3,5,10$ & $2,3,4$ \\

AutoFLSat & Intra/Inter Sat Links & EUROSAT & $1,3,5,10$ &$2,3,4$ \\

\bottomrule
\end{tabular}
}
\caption{\textit{AutoFLSat} specific parameter configuration settings, testing differences in \# of clusters, datasets, and epochs of training in each round. Three datasets are tested against, FEMINIST, CIFAR-10, and EUROSAT, with \#s of clusters in the constellation ranging from 2, 3, and 4. Varying epochs suitable for each type of training dataset are tested as well. There are 10 satellites per cluster.}
\label{tab:autoflConfigs}
\end{table*}

\subsection{Reading Heatmaps}
\label{app:readingheatmaps}
\begin{figure*}[ht]
\begin{minipage}{\textwidth}
    \begin{subfigure}{\textwidth}
    \centering
    \includegraphics[trim={7cm 0 7cm 0},clip,scale = 0.25]{figures/acc/fedAvgSat_acc.png}
    \label{fedavg:acc}
    \end{subfigure}

    \begin{subfigure}{\textwidth}
    \centering
    \includegraphics[trim={7cm 0 7cm 0},clip,scale = 0.25]{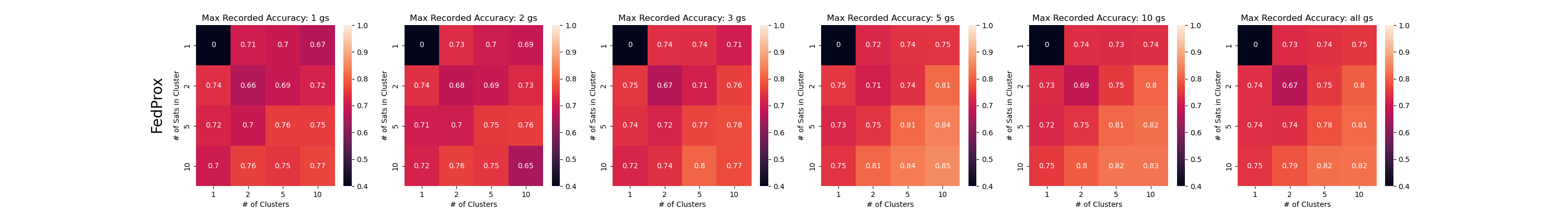}
    \label{fedprox:acc}
    \end{subfigure}
    
    \begin{subfigure}{\textwidth}
    \centering
    \includegraphics[trim={7cm 0 7cm 0},clip,scale = 0.25]{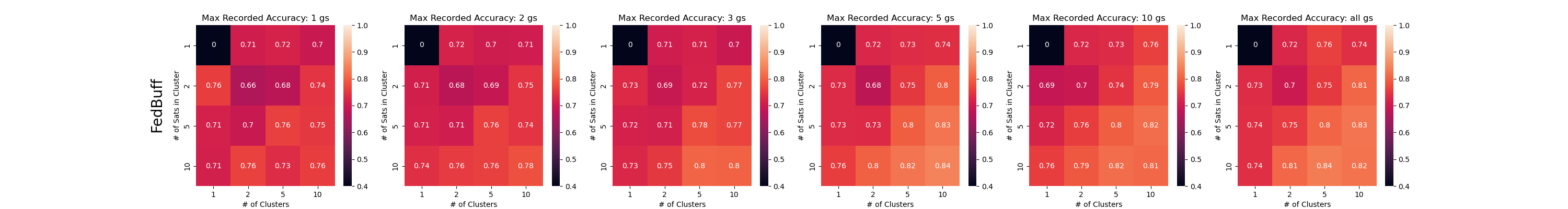}
    \caption{Baseline FL Algorithms, with no Scheduling or Intra Satellite Communications}
    \label{fedBuff:acc}
    \end{subfigure}
    
    \begin{subfigure}{\textwidth}
    \centering
    \includegraphics[trim={7cm 0 7cm 0},clip,scale = 0.25]{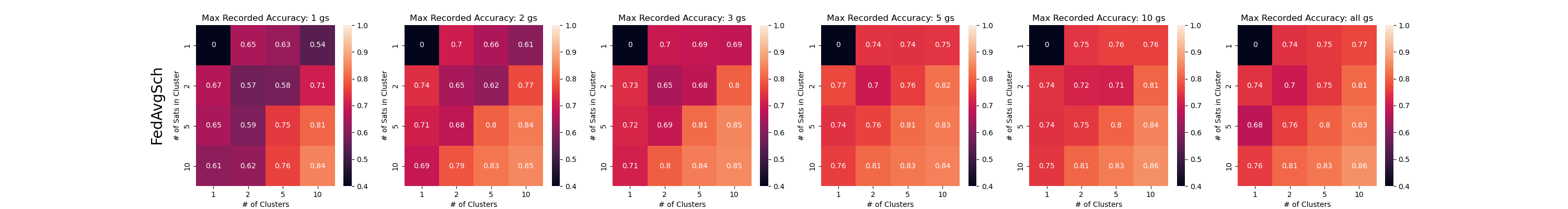}
    \label{fedavg2:acc}
    \end{subfigure}

    \begin{subfigure}{\textwidth}
    \centering
    \includegraphics[trim={7cm 0 7cm 0},clip,scale = 0.25]{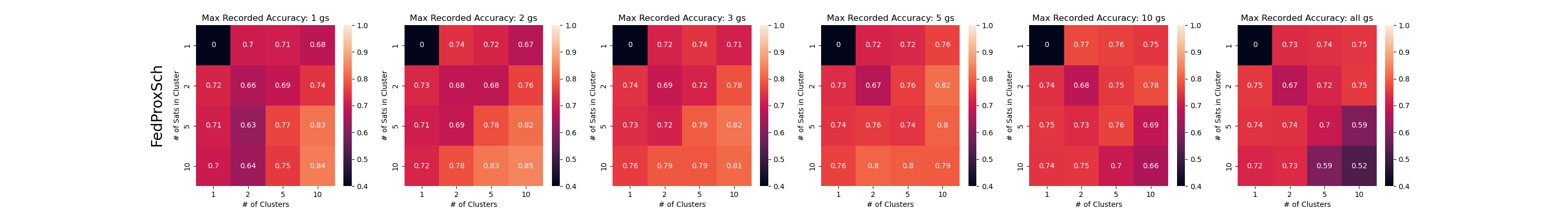}
    \label{fedprox2:acc}
    \end{subfigure}
    
    \begin{subfigure}{\textwidth}
    \centering
    \includegraphics[trim={7cm 0 7cm 0},clip,scale = 0.25]{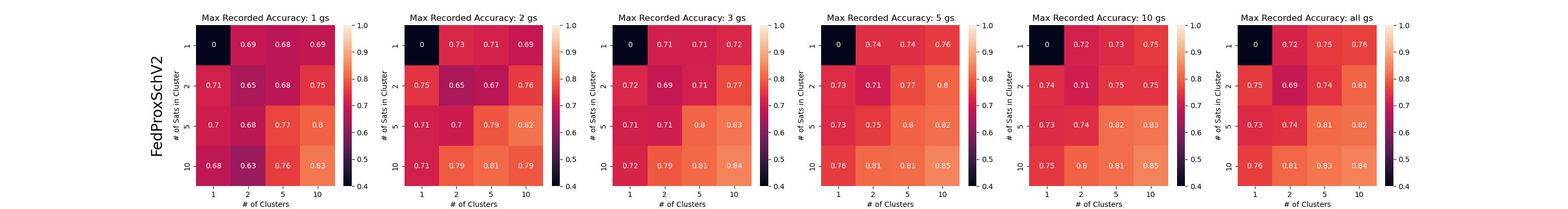}
    \caption{FL Algorithms with a Scheduler, but without Intra Satellite Communications}
    \label{fedprox22:acc}
    \end{subfigure}

    \begin{subfigure}{\textwidth}
    \centering
    \includegraphics[trim={0cm 0 0cm 0},clip,scale = 0.25]{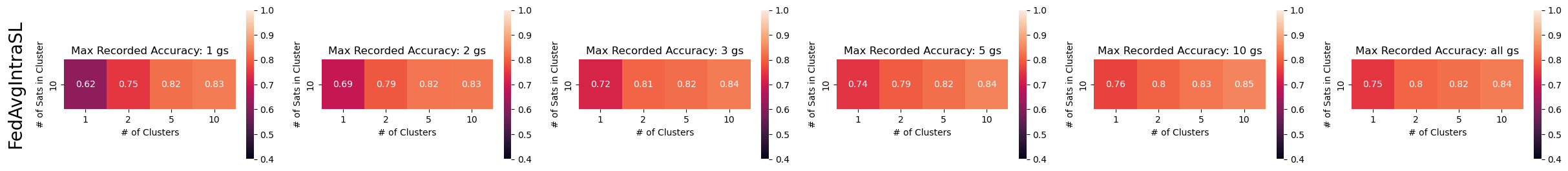}
    \label{fedavg3:acc}
    \end{subfigure}

    \begin{subfigure}{\textwidth}
    \centering
    \includegraphics[trim={.33cm 0 0cm 0},clip,scale = 0.25]{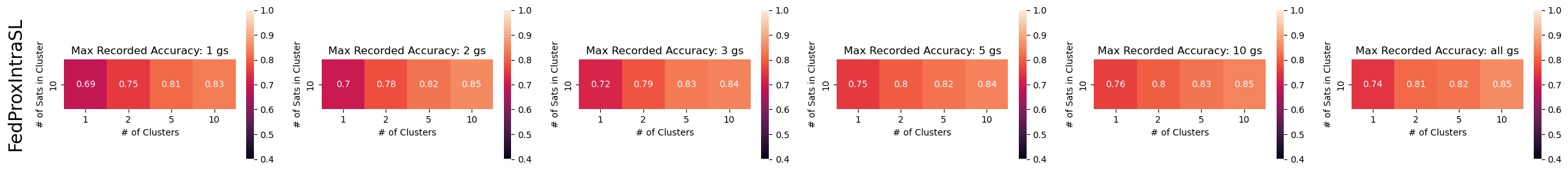}
    \caption{FL Algorithms with both a Scheduler, and enabled Intra Satellite Communications}
    \label{fedprox3:acc}
    \end{subfigure}

    \caption{Heatmaps depicting the maximum accuracy reached in the training of each satellite simulations, testing configuration parameters of varying \#s of clusters, \#s of satellites per cluster, and \#s of ground stations available to connect in the network. Multiple FL algorithms were implemented and tested, specifically \textit{FedAvg}, \textit{FedProx}, and \textit{FedBuff}, and measured against versions of the same algorithm but with scheduled and Intra SL enabled communications. All algorithms if provided enough aggregation opportunities could reach more than 80\% of accuracy. This was typically possible by optimizing for existing access windows through scheduling, or through the addition of more communication points through larger ground station networks or ISL enabled communications.}\label{fig:accs}
\end{minipage}
\end{figure*}

There are many parameters at play within the heatmap distributions outlined in \cref{fig:accs}, but the most important considerations to keep in mind while viewing the information are broken down into the following. 

Each individual square in each heatmap represents a certain type of satellite constellation setup, tested against a specific FL algorithm. Although difficult to see, the accuracy value for each of these individual squares, are written inside each location and are set between 0.4 and 1. The top left corner of every heat map is set to 0 because only one satellite client would be tested, which is not sufficient to perform FL operations.

Focusing in on each individual heatmap, the rows are representative of the number of satellites within each cluster (1, 2, 5, 10), and the columns represent the number of clusters (1, 2, 5, 10) within the constellation. 

Combining the two pieces of information, you get the number of satellites simulated for that experiment. For example, the experiment on the bottom right corner would be 100 total satellites from 10 clusters with 10 satellites on each cluster, while the top right color would be 10 total satellites from having 10 clusters with 1 satellite on each cluster.

Looking at the figure from a holistic point of view, each row represents the specific FL algorithm tested on the various satellite constellation parameters. The columns represent the number of ground stations available for communication (1, 2, 3, 5, 10, 13) for each batch of simulations.

\section{Additional Results}

\subsection{Duration Violin Plots}

\begin{figure}[htp]
    \centering
    \includegraphics[width=\linewidth]{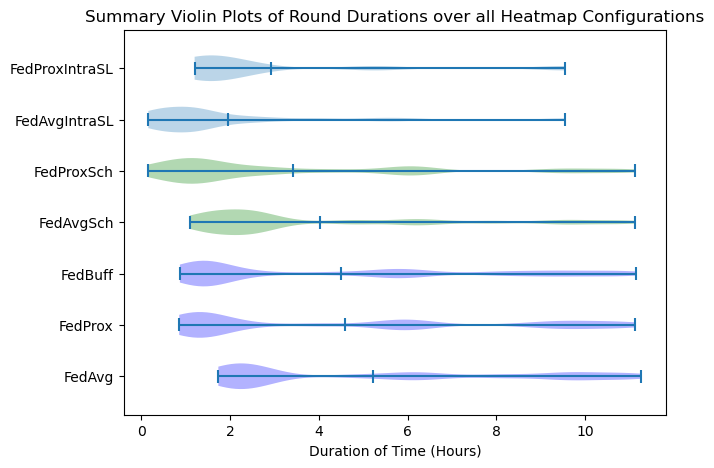}
    \caption{Summary violin plots of all heatmap values of FL round duration lengths, with three tickmarks indicating: (leftmost) minimum, (middle) average, and (rightmost) maximum values found in each heatmap group. Clear benefits in both the scheduling and ISL enabled communications can be found compared to the base methodologies, as mean FL round durations decrease by more than an hour in some algorithms.}
    \label{fig:summaryDuratinos}
\end{figure}

\begin{figure*}[htp]
    \centering
    \includegraphics[width=\textwidth]{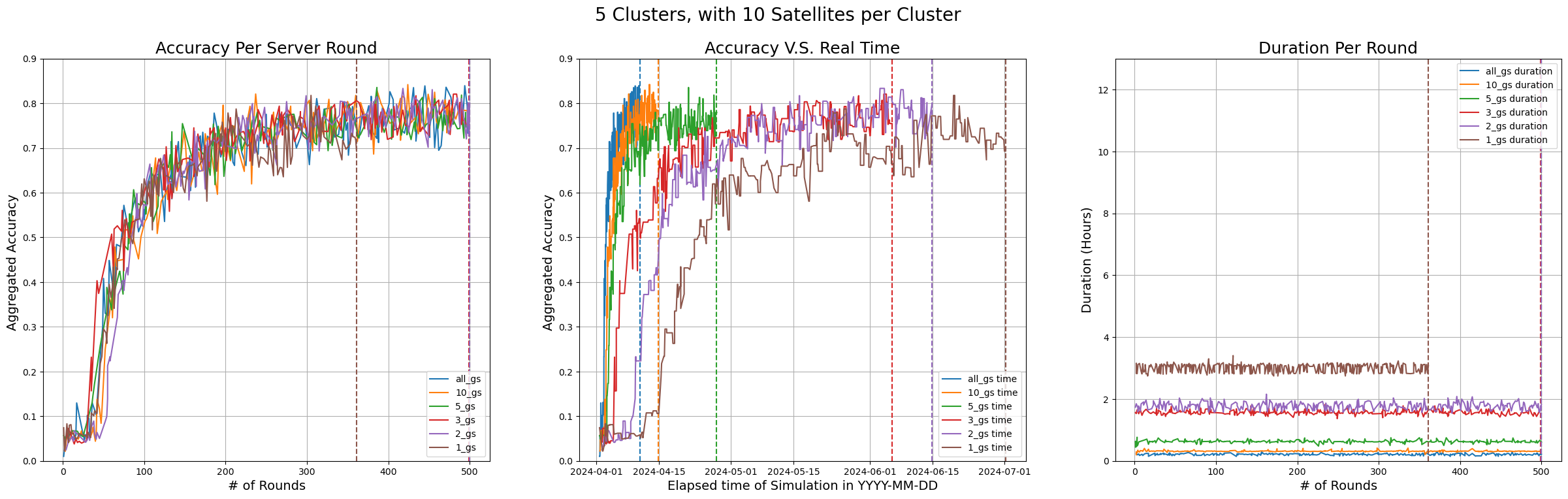}

    \caption{Performance of \textit{FedAvgSch} on a satellite constellation with 5 clusters, and 10 satellites per cluster. (Left) The first plot depicts accuracy v.s. the number of server rounds completed in the scenario, and almost all ground station networks are able to reach close to the 500 server round mark within the allotted 3 month simulation period. (Center) The second plot dives a bit deeper into the simulation time plotted against accuracy, and clear benefits of scheduling can be seen with the reduction of total simulation time dropping to 10 days for the largest ground station network. (Right) A final plot of duration times for each round is shown, with each simulation has less varying round aggregation times through the use of scheduling.}
    \label{fig:acc_line_fedAvg2}
\end{figure*}

\section{\textit{AutoFLSat}: Effects of Clusters and Epochs per Round on Performance Metrics}
\label{sec:clustertests}

To understand the effect of the different hierarchical training setups in \textit{AutoFLSat}, a parameter sweep is conducted varying the \# of clusters and epochs allowed for training. A table of these results are displayed in \cref{tab:autoFLSat}, where satellite constellation setups of 2, 3, and 4 clusters are tested against varying epoch per round of 20, 50, 100.

Comparing the evaluation metrics across each cluster scenario, it is clear that all cluster scenarios are able to reach a certain performance accuracy guarantee over 80\%. However, there are slight accuracy drops between the different cluster simulations, with slightly lower accuracies corresponding to smaller \#s of clusters in the satellite constellation. There may be several reasons for this discrepancy. In terms of satellite client selection distributions, it is noted that the design of \textit{AutoFLSat} leads to including every single satellite in each FL aggregation round. This reduces idle time significantly across all experiments, as both Intra and Inter Satellite Links are constantly active and sending model parameters across clients. However, this means performance differences in model accuracy may arise with smaller constellations. Less clients in a constellation corresponds to less data overall available to train for the global model, and less data for training typically leads to lower model accuracy performance. A satellite constellation with 2 clusters and 10 satellites per cluster only has half the clients (20) of a similar constellation with 4 clusters and 10 satellites per cluster (40), which could result in noticeable differences in performance.

In \cref{tab:autoFLSat}, the increase in the \#s of clusters in a constellation also seems to correspond to an increase in idle time and round duration. With additional clusters, the aggregation time for sending each cluster's global model to all other clusters is bound to increase, as more communication passes are needed to receive all other cluster models. If $C$ is the number of clusters in a constellation, each constellation will need $\frac{(C-1)C}{2}$ number of passes to ensure all cluster models have properly shared their parameters to all other clusters. Within the simulations tested, the increase in time duration is not immensely striking, but it is possible that with more clusters, the time could grow at a faster rate. Additional exploration into the dynamics of autonomy in larger mega constellations will need to be analyzed to understand this further.


\setlength{\tabcolsep}{30pt}

\begin{table*}[htp]
  \setlength{\tabcolsep}{0.3\tabcolsep}
\centering
\resizebox{0.9\linewidth}{!}{%
\begin{tabular}{cc|cccc}
\toprule
\textbf{\# of Clusters} & \textbf{\# of Epochs} & \textbf{Accuracy} & \textbf{Round Duration} & \textbf{Idle Time}& \textbf{Total Training Time} \\ \midrule
4 & 20 & 83.3\% & 12.1 min & 10.3 min& 20.6 Hours \\ 
4 & 50 & 83.01\%  & 11.93 min & 10.2 min & 21.28 Hours \\
4 & 100 & 82.5\%  & 12.3 min & 10.32 min & 28.1 Hours \\ \midrule
3 & 20 & 82.8\%  & 9.53 min & 8.4 min & 28 Hours\\
3 & 50 & 81.4\% & 9.4 min & 8.3 min & 22 Hours\\
3 & 100 & 82.2\%  & 9.3 min & 8.4 min & 18.79 Hours\\ \midrule
2 & 20 & 82.1\% & 4.5 min & 3.26 min & 14.3 Hours\\
2 & 50 & 82.5 \% & 4.6 min & 3.01 min & 15.2 Hours\\
2 & 100 & 81\% & 4.6 min & 2.6 min & 20.99 Hours\\

\bottomrule
\end{tabular}
}
\caption{Experiments of \textit{AutoFLSat} trained on the FEMNIST dataset, varying both the \#s of clusters and \#s of epochs tested. Values of the accuracy, round duration, idle time, and total training time are outlined. With increases in the \#s of clusters in the constellation, there is definitely an increase in idle time spent trying to accumulate every cluster's global model. These increases in idle time take up a large portion of the round duration, which may scale with the size of the constellation. The accuracy is able to be met at a reasonable time frame for each constellation setup, but a clear connection with accuracy percentages and constellation setups are not established.}
\label{tab:autoFLSat}
\end{table*}

\setlength{\tabcolsep}{30pt}

\begin{table*}[htp]
  \setlength{\tabcolsep}{0.3\tabcolsep}
\centering
\resizebox{0.75\linewidth}{!}{%
\begin{tabular}{c|ccccc}
\toprule
\textbf{\# of Clusters} & \textbf{Accuracy} & \textbf{Round Duration}& \textbf{Idle Time}& \textbf{Total Training Time} \\ \midrule
4 & 83.5\% & 12.4 min & 10.23 min & 14.1 Hours \\ \midrule
3 & 80.5\% & 9.6 min & 8.34 min & 11.2 Hours\\\midrule
2 & 82.1\% & 4.5 min & 2.6 min & 6.4 Hours\\

\bottomrule
\end{tabular}
}

\vspace*{3mm}
\caption{Experiments of \textit{AutoFLSat} trained on the EUROSAT dataset, varying both the \#s of clusters and \#s of epochs tested. Values of the accuracy, \# of rounds trained on, round duration, idle time, and total training time are outlined. Similar increases in idle time and round duration is found from the increase of \# of clusters as found in the FEMNIST dataset. However, speedy convergence is reached for image classification training on the EUROSAT satellite image patch collection, which gives promising indications of \texttt{AutoFLSat} providing proper convergence when trained on realistic data.}
\label{tab:eurosatautoflsat}
\end{table*}

\clearpage

\begin{figure*}[ht]
    \begin{subfigure}{\textwidth}
    \centering
    \includegraphics[trim={7cm 0 7cm 0},clip,scale = 0.25]{figures/dur/fedAvgSat_10_duration.png}
    \label{fedavg10:duration}
    \end{subfigure}

    \begin{subfigure}{\textwidth}
    \centering
    \includegraphics[trim={7cm 0 7cm 0},clip,scale = 0.25]{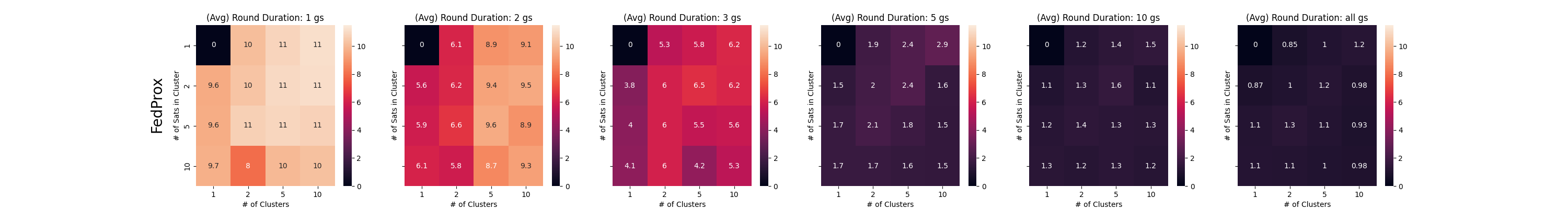}
    \label{fedProx:duration}
    \end{subfigure}

    \begin{subfigure}{\textwidth}
    \centering
    \includegraphics[trim={7cm 0 7cm 0},clip,scale = 0.25]{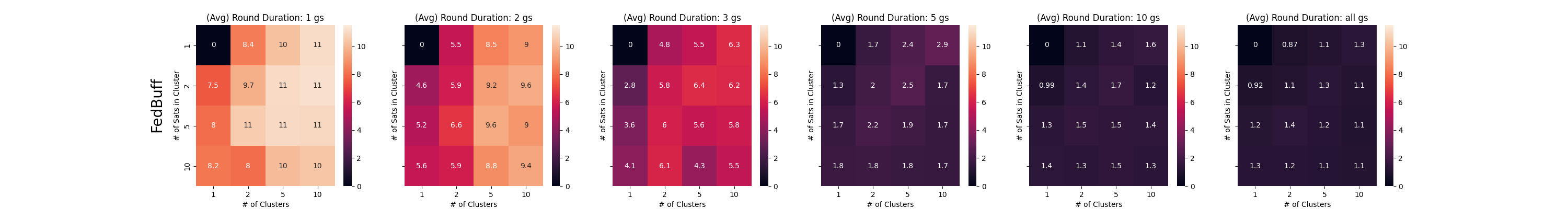}
    \caption{Baseline FL Algorithms, with no Scheduling or Intra Satellite Communications}
    \label{baseline:duration}
    \end{subfigure}

    \begin{subfigure}{\textwidth}
    \centering
    \includegraphics[trim={7cm 0 7cm 0},clip,scale = 0.25]{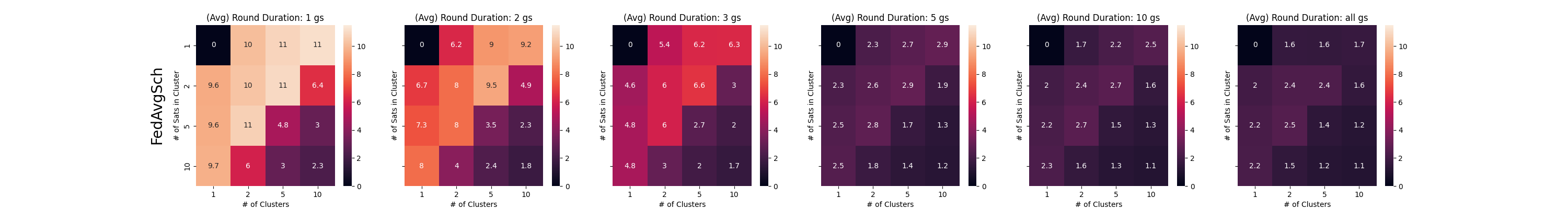}
    \label{fedavg2_10:duration}
    \end{subfigure}

    \begin{subfigure}{\textwidth}
    \centering
    \includegraphics[trim={7cm 0 7cm 0},clip,scale = 0.25]{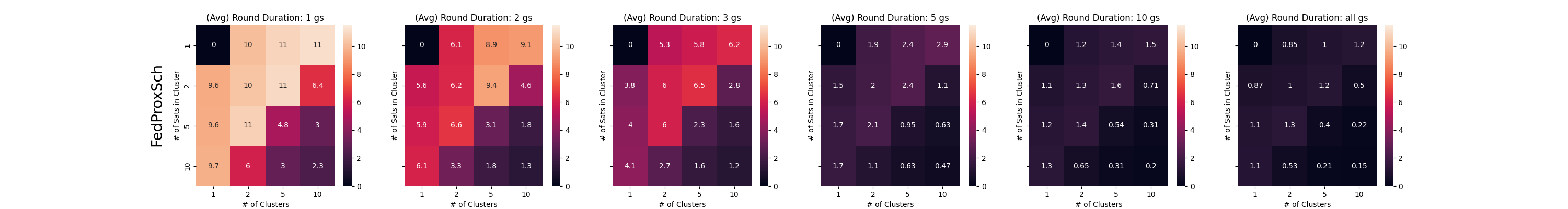}
    \label{fedProx2:duration}
    \end{subfigure}

    \begin{subfigure}{\textwidth}
    \centering
    \includegraphics[trim={7cm 0 7cm 0},clip,scale = 0.25]{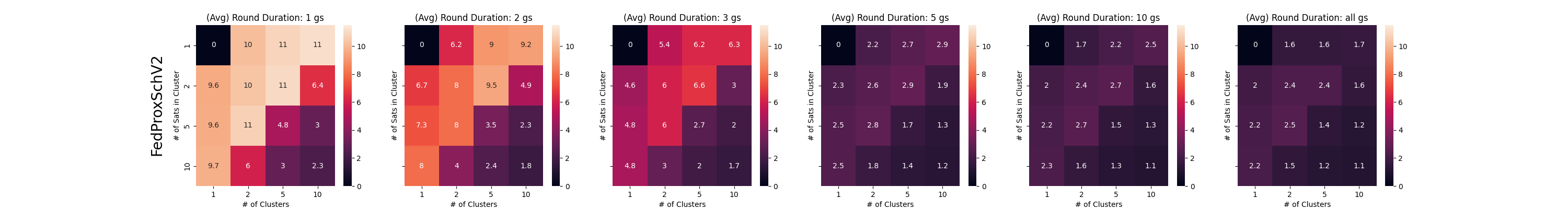}
    \caption{FL Algorithms with a Scheduler, but without Intra Satellite Communications}
    \label{fedBuff2
:duration}
    \end{subfigure}
    
    \begin{subfigure}{\textwidth}
    \centering
    \includegraphics[trim={0cm 0 0cm 0},clip,scale = 0.25]{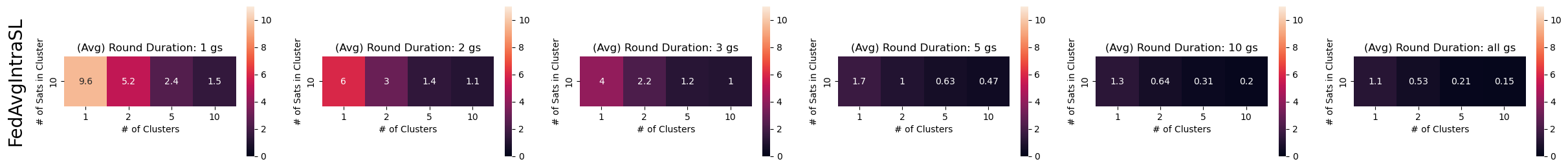}
    \label{fedavg3:dur}
    \end{subfigure}

    \begin{subfigure}{\textwidth}
    \centering
    \includegraphics[trim={.33cm 0 0cm 0},clip,scale = 0.25]{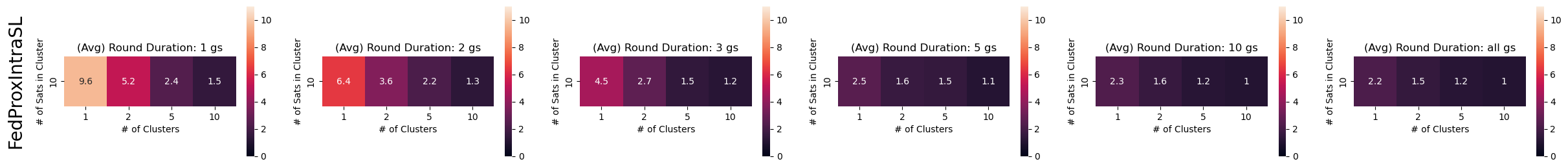}
    \caption{FL Algorithms with both a Scheduler, and enabled Intra Satellite Communications}
    \label{fedprox3:dur}
    
    \end{subfigure}
    \caption{Heatmaps depicting the average round duration per hour taken for FL training and aggregation, again testing changes in the \#s of clusters, \#s of satellites per cluster, and \#s of ground stations available to connect in the network. \textit{FedAvg}, \textit{FedProx}, and \textit{FedBuff}, are tested measured against versions of the same algorithm but with scheduled and Intra SL enabled communications. The number of ground stations seems to have the largest impact on round duration length, with clear drops in duration when testing 1, 2, 3, and 5 ground stations. However, a plateauing effect is seen with any additional ground stations added, suggesting a optimal ratio of satellite to ground stations in a constellation network.}\label{fig:base_durations}
\end{figure*}

\begin{figure*}[!ht]
    \begin{subfigure}{\textwidth}
    \centering
    \includegraphics[trim={7cm 0 7cm 0},clip,scale = 0.25]{figures/idle/fedAvgSat_10_iavg.png}
    \label{fedavg10:iavg}
    \end{subfigure}

    \begin{subfigure}{\textwidth}
    \centering
    \includegraphics[trim={7cm 0 7cm 0},clip,scale = 0.25]{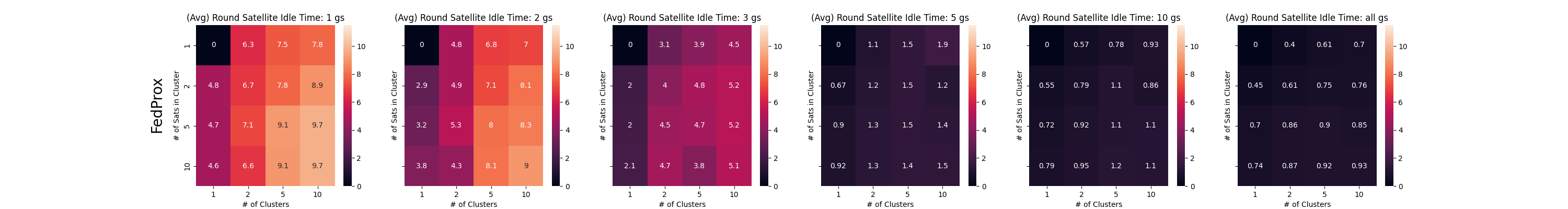}
    \label{fedProx:iavg}
    \end{subfigure}

    \begin{subfigure}{\textwidth}
    \centering
    \includegraphics[trim={7cm 0 7cm 0},clip,scale = 0.25]{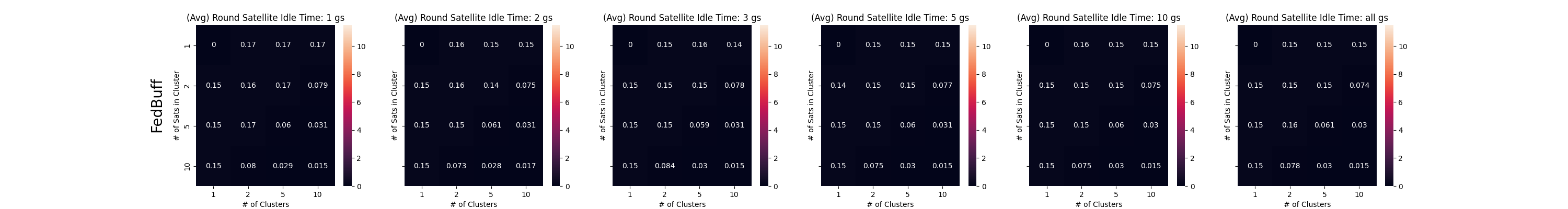}
    \caption{Baseline FL Algorithms, with no Scheduling or Intra Satellite Communications}
    \label{fedBuff:iavg}
    \end{subfigure}

    \begin{subfigure}{\textwidth}
    \centering
    \includegraphics[trim={7cm 0 7cm 0},clip,scale = 0.25]{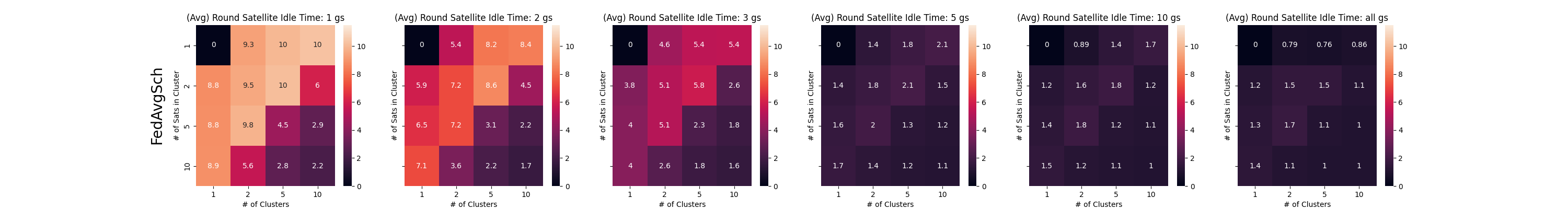}
    \label{fedavg2_10:iavg}
    \end{subfigure}

    \begin{subfigure}{\textwidth}
    \centering
    \includegraphics[trim={7cm 0 7cm 0},clip,scale = 0.25]{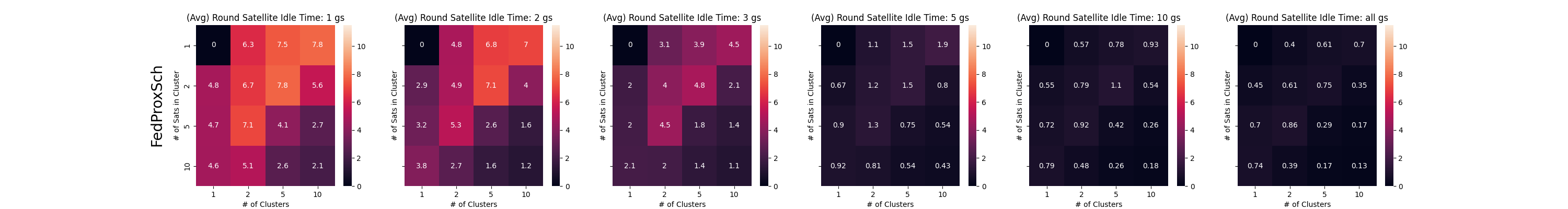}
    \label{fedProx2:iavg}
    \end{subfigure}

    \begin{subfigure}{\textwidth}
    \centering
    \includegraphics[trim={7cm 0 7cm 0},clip,scale = 0.25]{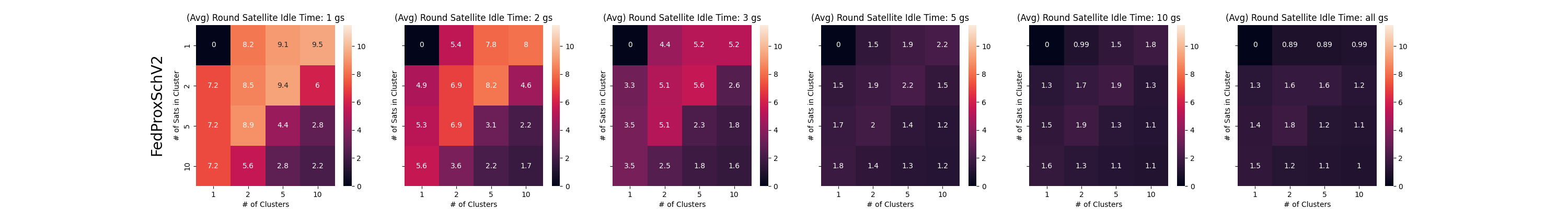}
    \caption{FL Algorithms with a Scheduler, but without Intra Satellite Communications}
    \label{fedBuff2
:iavg}
    \end{subfigure}
    
    \begin{subfigure}{\textwidth}
    \centering
    \includegraphics[trim={0cm 0 0cm 0},clip,scale = 0.25]{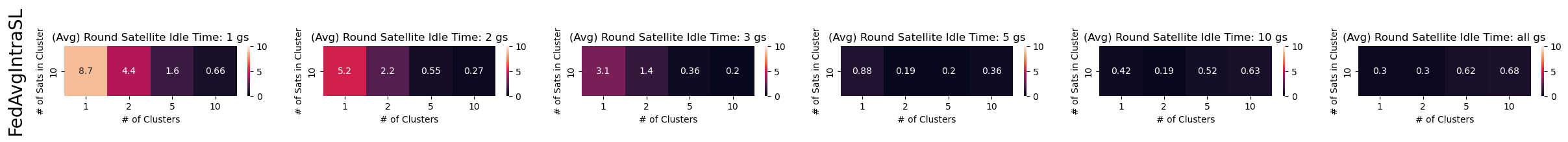}
    \label{fedavg3:idle}
    \end{subfigure}

    \begin{subfigure}{\textwidth}
    \centering
    \includegraphics[trim={.33cm 0 0cm 0},clip,scale = 0.25]{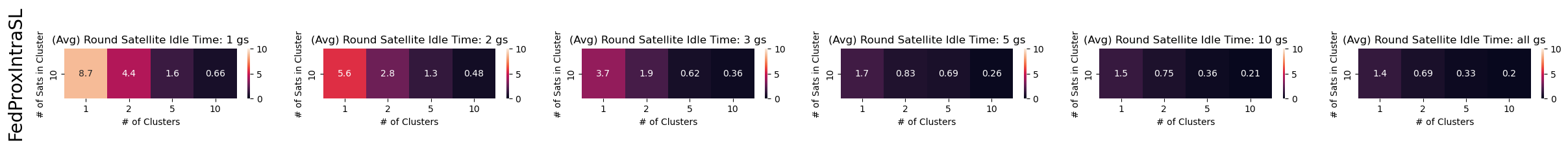}
    \caption{FL Algorithms with both a Scheduler, and enabled Intra Satellite Communications}
    \label{fedprox3:idle}
    \end{subfigure}
    
    \caption{Heatmaps depicting the satellite idle time per hour during round aggregation, testing changes in the \#s of clusters, \#s of satellites per cluster, and \#s of ground stations available to connect in the network. \textit{FedAvg}, \textit{FedProx}, and \textit{FedBuff}, are tested measured against versions of the same algorithm but with scheduled and Intra SL enabled communications. Again, the number of ground stations seems to have the largest impact on round duration length, but a point of interest is the drop in idle time through \textit{FedBuff}, which performs computation continously throughout each satellte's orbit.}\label{fig:idleavg_normal}
\end{figure*}

\end{document}